\documentclass[10pt,twocolumn,letterpaper]{article}

\usepackage[pagenumbers]{cvpr}      

\usepackage{pifont}%

\newcommand{\cmark}{\ding{51}}%
\newcommand{\xmark}{\ding{55}}%

\usepackage{graphicx}
\usepackage{amsmath}
\usepackage{amssymb}
\usepackage{booktabs}

\usepackage{amsmath,amsfonts}
\usepackage{algorithmic}
\usepackage{algorithm}
\usepackage{array}
\usepackage{textcomp}
\usepackage{stfloats}
\usepackage{url}
\usepackage{verbatim}
\usepackage{graphicx}
\usepackage{cite}
\usepackage{amsmath}
\usepackage{amssymb}
\usepackage[table]{xcolor}
\usepackage[pagebackref=true,breaklinks=true,colorlinks=true,bookmarks=false,linkcolor={red!50!black},urlcolor={darkgray!80!black},citecolor={green!50!black}]{hyperref}
\usepackage[group-separator={,}]{siunitx}
\usepackage{multirow}
\usepackage{tabularx}
\usepackage{booktabs} 
\usepackage{makecell}
\usepackage{graphbox}
\usepackage{footmisc}
\usepackage{bm}
\usepackage{caption}
\usepackage{arydshln}
\usepackage{dashrule}
\usepackage{subcaption}
\usepackage{enumitem}
\setlist[itemize]{noitemsep,nolistsep}

\usepackage[capitalize]{cleveref}
\crefname{section}{Sec.}{Secs.}
\Crefname{section}{Section}{Sections}
\Crefname{table}{Table}{Tables}
\crefname{table}{Tab.}{Tabs.}
\Crefname{figure}{Figure}{Figures}
\crefname{figure}{Fig.}{Figs.}
\Crefname{equation}{Equation}{Equations}
\crefname{equation}{Eq.}{Eqs.}

\usepackage{graphicx}
\usepackage{amsmath}
\usepackage{amssymb}
\usepackage{booktabs}

\usepackage{graphicx}
\usepackage{makecell}
\usepackage{comment}
\usepackage{amsmath,amssymb} 
\usepackage{color}
\usepackage{booktabs}
\usepackage{arydshln}
\usepackage{multirow}
\usepackage{hyperref}

\usepackage{caption}
\renewcommand{\paragraph}[1]{\noindent \textbf{#1}}

\newcommand{\blue}[1]{\textbf{\textcolor{mblue}{#1}}}
\newcommand{\rblue}[1]{{\textcolor{mblue}{#1}}}
\newcommand{\overall}[1]{\textbf{\textcolor{mgreenblue}{#1}}}
\newcommand{\nonecolor}[1]{}
\newcommand{\orange}[1]{\textbf{\textcolor{orange}{#1}}}
\newcommand{\oorange}[1]{{\textcolor{orange}{#1}}}
\newcommand{\red}[1]{\textcolor{red}{#1}}
\newcommand{\bred}[1]{\textbf{\textcolor{red}{#1}}}
\newcommand{\green}[1]{\textcolor{mgreen}{#1}}

\newcommand{\frag}[0]{{{\textit{fragments}}}}

\colorlet{lightcyan}{cyan!8}
\colorlet{lightpink}{pink!20}
\colorlet{lightgray}{gray!10}

\definecolor{mgray}{gray}{0.35}
\definecolor{mlgray}{gray}{0.85}

\definecolor{mred}{RGB}{238, 34, 12}
\definecolor{mgreen}{RGB}{1, 127, 0}
\definecolor{mblue}{RGB}{0, 77, 128}
\definecolor{mgreenblue}{RGB}{1,102,98}
\definecolor{orange}{RGB}{240, 120,0}

\usepackage[table]{xcolor}
\usepackage[pagebackref=true,breaklinks=true,colorlinks=true,bookmarks=false]{hyperref}

\usepackage[capitalize]{cleveref}
\crefname{section}{Sec.}{Secs.}
\Crefname{section}{Section}{Sections}
\Crefname{table}{Table}{Tables}
\crefname{table}{Tab.}{Tabs.}

\title{Exploring Video Quality Assessment on User Generated Contents 
 \\ from Aesthetic and Technical Perspectives }
\author{Haoning Wu\footnotemark[1]~$^{1}$ \qquad Erli Zhang\footnotemark[1]~$^{1}$ \qquad Liang Liao\footnotemark[1]~$^1$ \qquad Chaofeng Chen$^1$  \qquad \\ Jingwen Hou$^1$  \qquad  Annan Wang$^1$ \qquad Wenxiu Sun$^2$ \qquad Qiong Yan$^2$ \qquad Weisi Lin$^{1}$ \\
$^1$  S-Lab, Nanyang Technological University \qquad  $^2$ Sensetime Research and Tetras AI
}
\begin{document}

\maketitle

\begin{abstract}
The rapid increase in user-generated content (UGC) videos calls for the development of effective video quality assessment (VQA) algorithms. However, the objective of the UGC-VQA problem is still ambiguous and can be viewed from two perspectives: the \textbf{\textit{\green{technical perspective}}}, measuring the perception of distortions; and the \textbf{\textit{\blue{aesthetic perspective}}}, which relates to preference and recommendation on contents. To understand how these two perspectives affect overall subjective opinions in UGC-VQA, we conduct a large-scale subjective study to collect human quality opinions on the overall quality of videos as well as perceptions from aesthetic and technical perspectives. The collected Disentangled Video Quality Database (\textbf{DIVIDE-3k}) confirms that human quality opinions on UGC videos are universally and inevitably affected by both aesthetic and technical perspectives. In light of this, we propose the \underline{D}isentangled \underline{O}bjective \underline{V}ideo Quality \underline{E}valuato\underline{r} (\textbf{DOVER}) to learn the quality of UGC videos based on the two perspectives. The DOVER proves state-of-the-art performance in UGC-VQA under very high efficiency. With perspective opinions in DIVIDE-3k, we further propose \textbf{DOVER++}, the first approach to provide reliable clear-cut quality evaluations from a single aesthetic or technical perspective. Code at \small{\url{ https://github.com/VQAssessment/DOVER}}.

\end{abstract}

\begin{figure}
    \center  \includegraphics[width=0.976\linewidth]{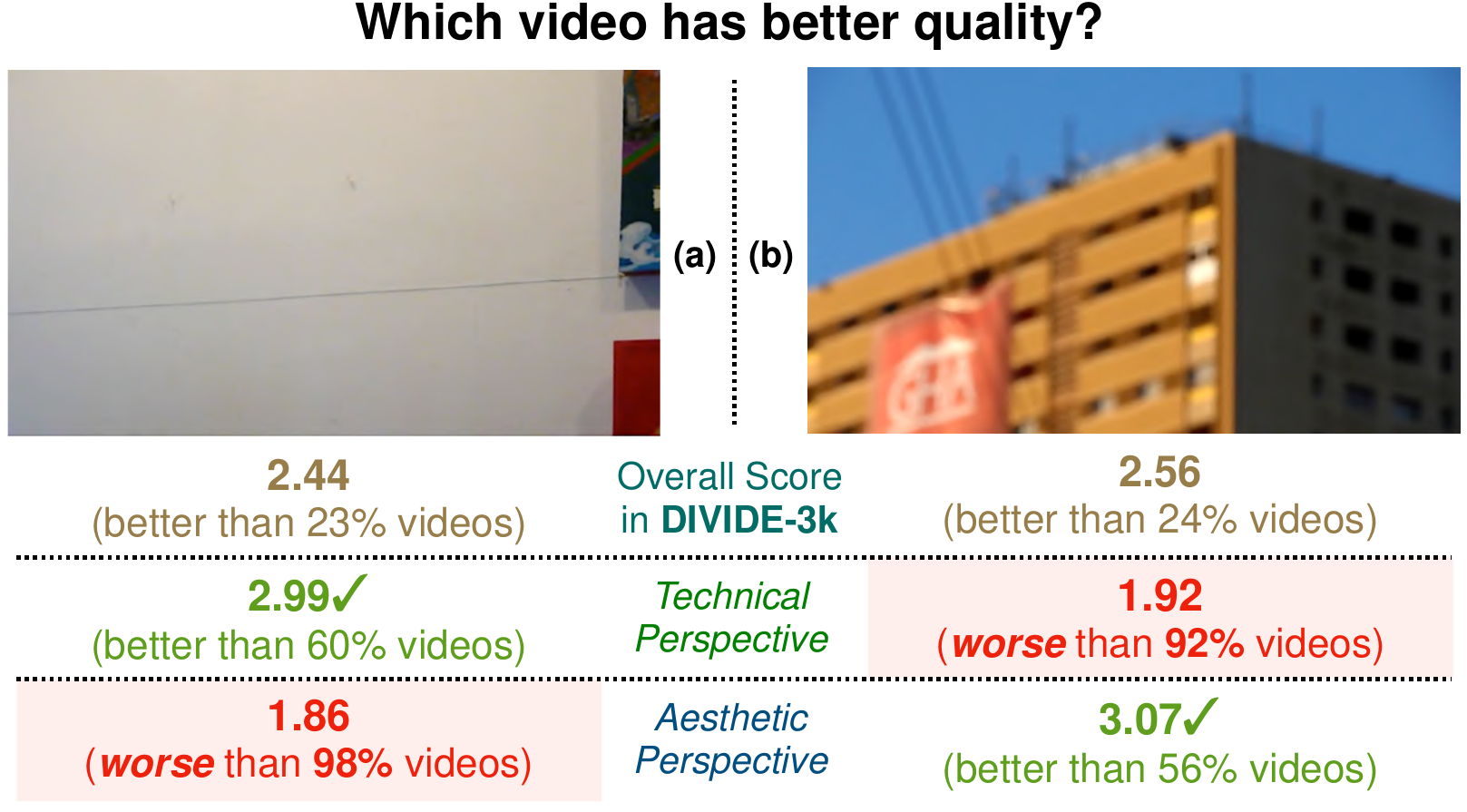}
    \vspace{-7pt}
    \captionof{figure}{\textbf{Which video has better quality}: a clear video with meaningless contents \textbf{{(a)}} or a blurry video with meaningful contents {\textbf{(b)}}? Viewing from different perspectives (\textbf{\textit{\blue{aesthetic}}}/\textbf{\textit{\green{technical}}}) may produce different judgments, motivating us to collect \textbf{DIVIDE-3k}, which is the first UGC-VQA dataset with opinions from multiple perspectives. More multi-perspective quality comparisons in our dataset are shown in \textcolor{brown}{\textit{supplementary \textbf{Sec. A}.}}}
    \vspace{-18pt}
    \label{fig:ugc_process}
\end{figure}%

\section{Introduction}
\label{sec:intro}


Understanding and predicting human quality of experience (QoE) on diverse in-the-wild videos has been a long-existing and unsolved problem. Recent Video Quality Assessment (VQA) studies have gathered enormous human quality opinions~\cite{pvq,mlsp,kv1k,vqc,ytugccc} on in-the-wild user-generated contents (UGC) and attempted to use machine algorithms~\cite{vsfa,rfugc,internetvqa} to learn and predict these opinions, known as the \textbf{UGC-VQA problem}~\cite{videval}. However, due to the diversity of contents in UGC videos and the lack of reference videos during subjective studies, these human-quality opinions are still ambiguous and may relate to different perspectives.

\renewcommand{\thefootnote}{\fnsymbol{footnote}}
\footnotetext[1]{The authors contribute equally to this paper.} 

Conventionally, VQA studies~\cite{videval,tlvqm,niqe,tpqi,fastvqa} are concerned with the \textit{\textbf{\green{technical perspective}}}, aiming at measuring distortions in videos (\textit{e.g.},~\textit{blurs, artifacts}) and their impact on quality, so as to compare and guide technical systems such as cameras~\cite{dxomark,spaq}, restoration algorithms~\cite{basicvsr,swinir,mmp} and compression standards~\cite{h264}. Under this perspective, the video with clear textures in Fig.~\ref{fig:ugc_process}(a) should have notably better quality than the blurry video in Fig.~\ref{fig:ugc_process}(b). On the other hand, several recent studies~\cite{sfa,vsfa,rfugc,mlsp,discovqa} notice that preferences on non-technical semantic factors (\textit{e.g.},~\textit{contents, composition}) also affect human quality assessment on UGC videos. Human experience on these factors is usually regarded as the \textit{\textbf{\blue{aesthetic perspective}}}~\cite{avaiaa,mlspiaa,distilliaa,racniaa,piaadataset,objiaa} of quality evaluation, which considers the video in Fig.~\ref{fig:ugc_process}(b) as better quality due to its more meaningful contents and is preferred for content recommendation systems on platforms such as YouTube or TikTok. However, how aesthetic preference plays the impact on final human quality opinions of UGC videos is still debatable~\cite{pvq,mlsp} and requires further validation. 

To investigate the impact of aesthetic and technical perspectives on human quality perception of UGC videos, we conduct the first comprehensive subjective study to collect opinions from both perspectives, as well as overall opinions on a large number of videos. 
We also conduct subjective reasoning studies to explicitly gather information on how much each individual's overall quality opinion is influenced by aesthetic and technical perspectives. With overall 450K opinions on 3,590 diverse UGC videos, we construct the first \underline{Di}sentangled \underline{Vi}deo Quality \underline{D}atabas\underline{e} (\textbf{DIVIDE-3k}). After calibrating our study on the DIVIDE-3k with existing UGC-VQA subjective studies, we observe that human quality perception on UGC videos is broadly and inevitably \textit{\textbf{affected by both aesthetic and technical perspectives}}. As a consequence, the overall subjective quality scores between the two videos in Fig.~\ref{fig:ugc_process} with different qualities from either one of the two perspectives could be similar.

Motivated by the observation from our subjective study, we aim to develop an objective UGC-VQA method that accounts for both aesthetic and technical perspectives. To achieve this, we design the View Decomposition strategy, which divides and conquers aesthetic-related and technical-related information in videos, and propose the \underline{D}isentangled \underline{O}bjective \underline{V}ideo Quality \underline{E}valuato\underline{r} (\textbf{DOVER}).  DOVER consists of two branches, each dedicated to focusing on the effects of one perspective. Specifically, based on the different characteristics of quality issues related to each perspective, we carefully design inductive biases for each branch, including \textit{specific inputs, regularization strategies, and pre-training}. The two branches are supervised by the overall scores (affected by both perspectives) to adapt for existing UGC-VQA datasets~\cite{vqc,kv1k,pvq,cvd,qualcomm,ytugc}, and additionally supervised by aesthetic and technical opinions exclusively in the \textbf{DIVIDE-3k} (denoted as \textbf{DOVER++}). Finally, we obtain the overall quality prediction via a subjectively-inspired fusion of the predictions from the two perspectives. With the subjectively-inspired design, the proposed DOVER and DOVER++ not only reach better accuracy on the overall quality prediction but also provide more reliable quality prediction from aesthetic and technical perspectives, catering for practical scenarios.

Our contributions can be summarized as four-fold:

\begin{enumerate} [topsep=0pt,itemsep=2pt,parsep=0pt]
\renewcommand{\labelenumi}{\theenumi)}
\item We collect the \textbf{DIVIDE-3k} (3,590 videos), the first UGC-VQA database that contains 450,000 subjective quality opinions from aesthetic and technical perspectives as well as their effects on overall quality scores.

\item By analyzing opinions, we observe that human quality perception is broadly affected by both aesthetic and technical perspectives in the UGC-VQA problem, better explaining the human perceptual mechanism on it.

\item We propose the \textbf{DOVER}, a subjectively-inspired video quality evaluator with two branches focusing on aesthetic and technical perspectives. The DOVER demonstrates state-of-the-arts on the \textbf{all} UGC-VQA datasets.

\item Our methods can provide quality predictions from a single perspective, which can be applied as metrics for camera systems (\textbf{\green{\textit{technical}}}) or content recommendation (\blue{\textit{aesthetic}}), or for personalized VQA~(Sec.~\ref{sec:personalized}).



    
\end{enumerate}

\section{Related Works}



\paragraph{Databases and Subjective Studies on UGC-VQA.} 
 Unlike traditional VQA databases \cite{livevqa,csiqvqa,cvd,qualcomm}, UGC-VQA databases~\cite{pvq,vqc,kv1k,ytugccc} directly collect from real-world videos from direct photography, YFCC-100M~\cite{yfcc} database or YouTube~\cite{ytugc} videos. With each video having unique content and being produced by either professional or non-professional users \cite{rfugc,internetvqa}, quality assessment of UGC videos can be more challenging and less clear-cut compared to traditional VQA tasks. Additionally, the subjective studies in UGC-VQA datasets are usually carried out by crowdsourced users \cite{crowdsource} with no reference videos. These factors may lead to the ambiguity of subjective quality opinions in UGC-VQA which can be affected by different perspectives.

\paragraph{Objective Methods for UGC-VQA.}
Classical VQA methods \cite{niqe,bofqa,rrstedqa,diivine,stgreed,vmaf,tlvqm,videval,brisque,viideo,vbliinds,tpqi} employ handcrafted features to evaluate video quality. However, they do not take the effects of semantics into consideration, resulting in reduced accuracy on UGC videos.
Noticing that UGC-VQA is deeply affected by semantics, deep VQA methods \cite{fastvqa, dctqa, cnn+lstm,deepvqa, gstvqa, vsfa, mlsp, rirnet, dstsvqa, svqa, mdtvsfa} are becoming predominant in this problem. For instance, VSFA~\cite{vsfa} conducts subjective studies to demonstrate videos with attractive content receive higher subjective scores. Therefore, it uses the semantic-pretrained  ResNet-50~\cite{he2016residual} features instead of handcrafted features, followed by plenty of recent works~\cite{lsctphiq,pvq,cnntlvqm,mlsp,fastvqa,discovqa} that improve the performance for UGC-VQA. However, these methods, which are directly driven by ambiguous subjective opinions, can hardly explain what factors are considered in their quality predictions, hindering them from providing reliable and explainable quality evaluations on real-world scenarios (\textit{e.g.}, distortion metrics and recommendations).

\section{The DIVIDE-3k Database}

\begin{figure}
    \centering
    \includegraphics[width=\linewidth]{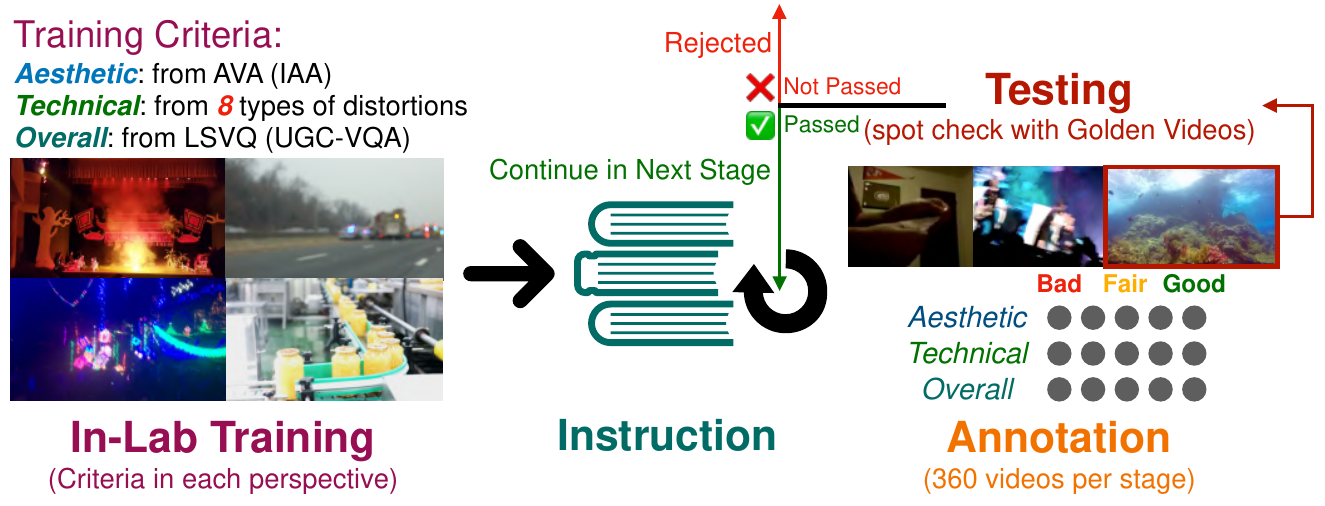}\vspace{-10pt}    \caption{The in-lab subjective study on videos in \textbf{DIVIDE-3k}, including Training, Instruction, Annotation and Testing, discussed in Sec.~\ref{sec:subprocess}.}
    \label{fig:divideprocess}
    \vspace{-17pt}
\end{figure}

In this section, we introduce the proposed Disentangled Video Quality Database (\textbf{DIVIDE-3k}, Fig.~\ref{fig:divideprocess}), along with the multi-perspective subjective study. The database includes 3,590 UGC videos, on which we collected 450,000 human opinions. Different from other UGC-VQA databases~\cite{pvq,mlsp,kv1k}, the subjective study is conducted \bred{in-lab} to reduce the ambiguity of perspective opinions.

\subsection{Collection of Videos}

\paragraph{Sources of Videos.} The 3,590-video database is mainly collected from two sources: 1) the YFCC-100M~\cite{yfcc} social media database; 2) the Kinetics-400~\cite{k400data} video recognition database, collected from YouTube, which has in total 400,000 videos. Voices are removed from all videos. 

\paragraph{Getting the subset for annotation.} 
Similar to existing studies~\cite{kv1k,pvq}, we would like the sampled video database able to represent the overall quality of the original larger database. Therefore, we first histogram all 400,000 videos with spatial~\cite{niqe}, temporal~\cite{tpqi}, and semantic indices~\cite{clipiqa}. Then, we randomly select a subset of 3,270 videos from the 400,000 videos that match the histogram from the three dimensions~\cite{matchhistogram} as in~\cite{pvq,kv1k}. Several examples from DIVIDE-3k are provided in the supplementary. We also select 320 videos from the LSVQ~\cite{pvq}, the most recent UGC-VQA database, to examine the calibration between DIVIDE-3k and existing UGC-VQA subjective studies (see in Tab.~\ref{tab:ugccalibration}).

\subsection{In-lab Subjective Study on Videos}
\label{sec:subprocess}
To ensure a clear understanding of the two perspectives, we conduct in-lab subjective experiments instead of crowdsourced, with 35 trained annotators (including 19 male and 16 female) participating in the full annotation process of Training, Testing and Annotation. All videos are downloaded to local computers before annotation to avoid transmission errors. The main process of the subjective study is illustrated in Fig.~\ref{fig:divideprocess}, discussed step-by-step as follows. \textcolor{brown}{\textit{Extended details about the study are in supplementary \textbf{Sec. A}.}}

\paragraph{Training.} Before annotation, we provide clear criteria with abundant examples of the three quality ratings to train the annotators. For \textbf{\textit{\blue{aesthetic rating}}}, we select example images with {\green{\textit{good}}}, {\oorange{\textit{fair}}} and {\red{\textit{bad}}} aesthetic quality from the aesthetic assessment database AVA~\cite{avaiaa}, each for 20 images, as calibration for aesthetic evaluation. For \textit{\textbf{\green{technical rating}}}, we instruct subjects to rate purely based on technical distortions and provide 5 examples for each of the following eight common distortions: \textit{1) noises; 2) artifacts; 3) low sharpness; 4) out-of-focus; 5) motion blur; 6) stall; 7) jitter; 8) over/under-exposure.} For overall quality rating, we select 20 videos each with {\green{\textit{good}}}, {\oorange{\textit{fair}}} and {\red{\textit{bad}}} quality as examples, from the UGC-VQA dataset LSVQ~\cite{pvq}.

\begin{figure}
    \centering
    \includegraphics[width=\linewidth]{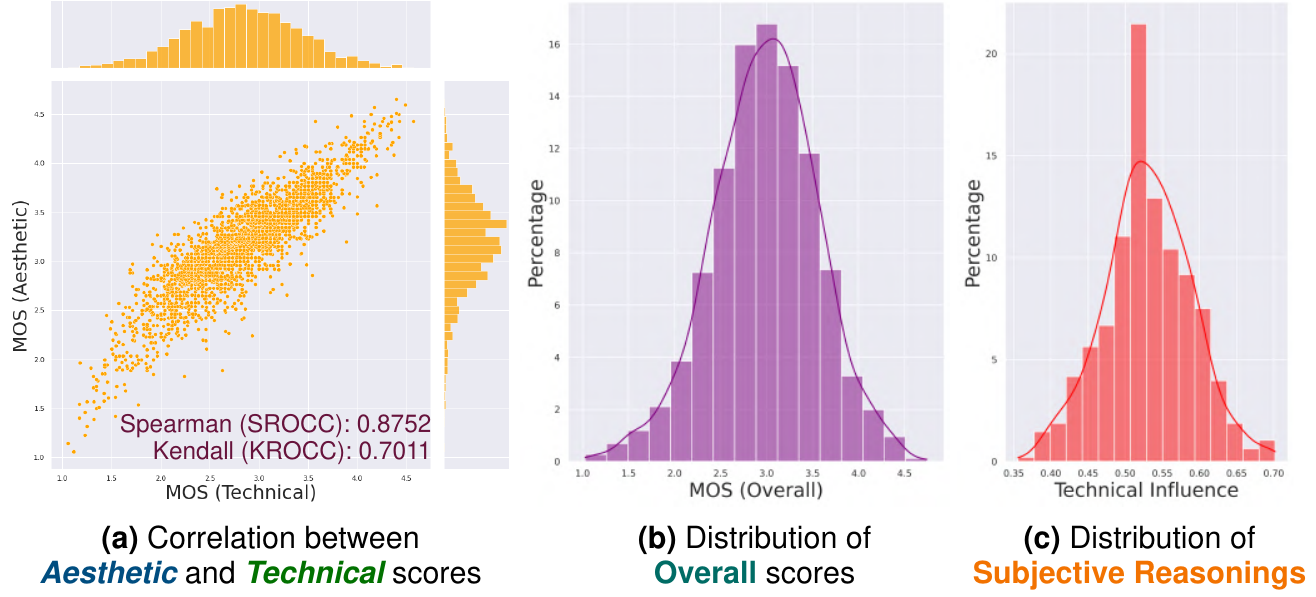}
    \vspace{-16pt}
    \caption{\textbf{Statistics in DIVIDE-3k}: \textbf{(a)} The correlations between aesthetic and technical perspectives, and distributions \textbf{(b)} of overall quality ($\mathrm{MOS}$) \& \textbf{(c)} subject-rated proportion of technical impact on overall quality.}
    \label{fig:histogram}
    \vspace{-10pt}
\end{figure}

\paragraph{During Experiment: Instruction and Annotation.} We divide the subjective experiments into 40 videos per group, and 9 groups per stage. Before each stage, we instruct the subjects on how to label each specific perspective:

\begin{itemize}
\item \textbf{\blue{Aesthetic Score}}: Please rate the video's quality based on aesthetic perspective (\textit{e.g.}, semantic preference).
\item \textbf{\green{Technical Score}}: Please rate the video's quality with only consideration of technical distortions.
\item \textbf{\overall{Overall Score}}: Please rate the quality of the video.
\item \textbf{\orange{Subjective Reasoning}}: Please rate how \underline{\textbf{your}} overall score is
 impacted by aesthetic or technical perspective.
\end{itemize}
Specifically, for the subjective reasoning, subjects need to rate the proportion of \textbf{\green{\textit{technical}}} impact in the overall score for each video among $[0, 0.25, 0.5, 0.75, 1]$, while rest proportion is considered as \textbf{\blue{\textit{aesthetic}}} impact.

\paragraph{Testing with Golden Videos.} For testing, we randomly insert 10 \textbf{golden videos} in each stage as a spot check to ensure the quality of annotation, and the subject will be rejected and not join the next stage if the annotations on the golden videos severely deviate from the standards.

\begin{figure*}
    \centering
    \includegraphics[width=1.00\linewidth]{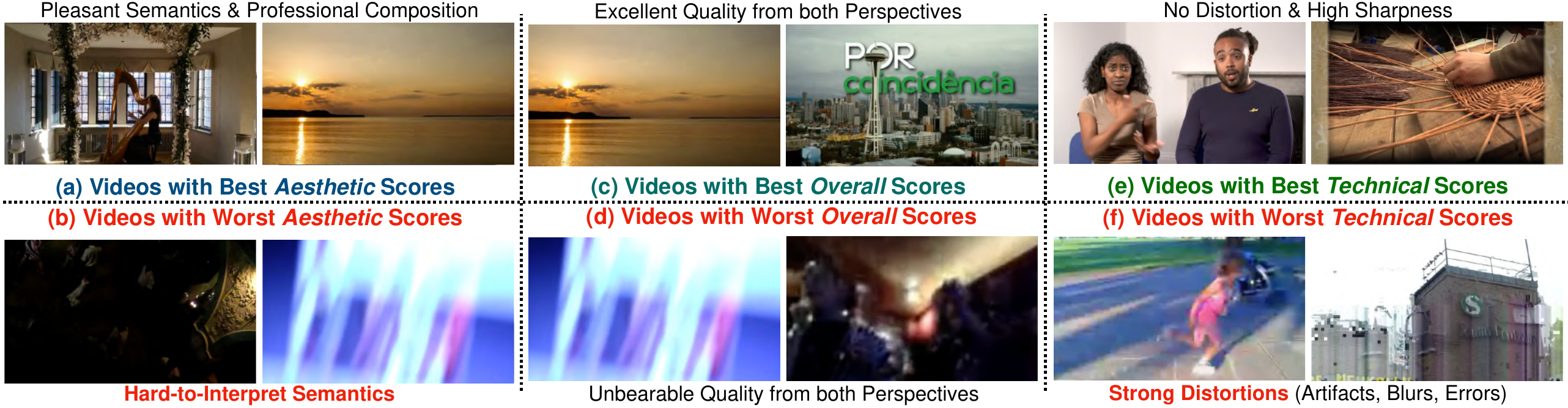}
    \vspace{-16pt}
    \caption{Videos with best and worst scores in aesthetic perspective, technical perspective and overall quality perception in the \textbf{DIVIDE-3k}. The aesthetic perspective is more concerned with semantics or composition of videos, while the technical perspective is more related to low-level textures and distortions.}
    \label{fig:dividebestworst}
    \vspace{-10pt}
\end{figure*}


\subsection{Observations}
\label{sec:3c}

\begin{table}
    \centering
    \footnotesize
    \renewcommand\arraystretch{1.3}
    \setlength\tabcolsep{5pt}
    \caption{\textbf{Effects of Perspectives}: {The correlations between different perspectives and overall quality~($\mathrm{MOS}$)} for all 3,590 videos in \textbf{DIVIDE-3k}.}
    \vspace{-8pt}
    \resizebox{\linewidth}{!}{\begin{tabular}{l|c|c|c|c}
        \hline
        Correlation to $\mathrm{MOS}$ & \blue{$\mathrm{MOS}_\mathrm{A}$} & \green{$\mathrm{MOS}_\mathrm{T}$} & $\mathrm{MOS}_\mathrm{A}+\mathrm{MOS}_\mathrm{T}$ & $0.428\mathrm{MOS}_\mathrm{A}+0.572\mathrm{MOS}_\mathrm{T}$\\ \hline
         Spearman (SROCC$\uparrow$) & 0.9350 & 0.9642 & 0.9827 & \bred{0.9834}  \\
         Kendall (KROCC$\uparrow$) & 0.7894 & 0.8455 & 0.8909 & \bred{0.8933} \\
         \hline
    \end{tabular}}
    \label{tab:subconclusion}
    \vspace{-10pt}
\end{table}

\begin{table}
    \centering
    \footnotesize
    \renewcommand\arraystretch{1.2}
    \setlength\tabcolsep{8pt}
    \caption{\textbf{Calibration with Existing}: The correlations of between different ratings in \textbf{DIVIDE-3k} and existing scores in LSVQ~\cite{pvq} ($\mathrm{MOS}_\text{existing}$).}
    \vspace{-8pt}
    \resizebox{\linewidth}{!}{\begin{tabular}{l|c|c|c|c}
        \hline
        Correlation to $\mathrm{MOS}_\text{existing}$ & \blue{$\mathrm{MOS}_\mathrm{A}$} & \green{$\mathrm{MOS}_\mathrm{T}$} & $\mathrm{MOS}_\mathrm{A}+\mathrm{MOS}_\mathrm{T}$ & $\mathrm{MOS}$\\ \hline
         Spearman (SROCC$\uparrow$) & 0.6956 & 0.7374 & 0.7632 & \bred{0.7680}  \\
         Kendall (KROCC$\uparrow$) & 0.5073 & 0.5469 & 0.5797 & \bred{0.5822} \\
         \hline
    \end{tabular}}
    \label{tab:ugccalibration}
    \vspace{-14pt}
\end{table}

\paragraph{Effects of Two Perspectives.} To validate the effects of two perspectives, we first quantitatively assess the correlation between the two perspectives and overall quality. Denote the mean aesthetic opinion as \blue{$\mathrm{MOS}_\mathrm{A}$}, mean technical opinion as \green{$\mathrm{MOS}_\mathrm{T}$}, mean overall opinion as $\mathrm{MOS}$, the Spearman and Kendall correlation between different perspectives are listed in Tab.~\ref{tab:subconclusion}. From Tab.~\ref{tab:subconclusion}, we notice that the weighted sum of both perspectives is a better approximation of overall quality than either single perspective. Consequently, methods~\cite{vsfa,bvqa2021,pvq} that naively regress from overall $\mathrm{MOS}$ might not provide pure technical quality predictions due to the inevitable effect of aesthetics. The best/worst videos (Fig.~\ref{fig:dividebestworst}) in each dimension also support this observation.

\paragraph{Calibration with Existing Study.} To validate whether the observation can be extended for existing UGC-VQA subjective studies, we select 320 videos from LSVQ~\cite{pvq} to compare quality opinions from multi-perspectives with existing scores of these videos. As shown in Tab.~\ref{tab:ugccalibration}, the overall quality score is more correlated with the existing score than scores from either perspective, further suggesting considering human quality opinion as a fusion of both perspectives might be a better approximation in the UGC-VQA problem.

\paragraph{Subjective Reasoning.} In the DIVIDE-3k, we conducted the first subjective reasoning study during the human quality assessment. Fig.~\ref{fig:histogram}(c) illustrates the mean technical impact for each video, ranging among $[0.364,0.698]$. The results of reasoning further explicitly validate our aforementioned observation, that human quality assessment is affected by opinions from both aesthetic and technical perspectives.

\section{The Approaches: DOVER and DOVER++}
\label{sec:4}

Observing that overall quality opinions are affected by both aesthetic and technical perspectives from subjective studies in \textbf{DIVIDE-3k}, we propose to distinguish and investigate the aesthetic and technical effects in a UGC-VQA model based on the View Decomposition strategy (Sec.~\ref{sec:viewdecomposition}). The proposed \underline{D}isentangled \underline{O}bjective \underline{V}ideo Quality \underline{E}valuato\underline{r} (\textbf{DOVER}) is built up with an aesthetic branch (Sec.~\ref{sec:aesbranch}) and a technical branch (Sec.~\ref{sec:tecbranch}). The two branches are separately supervised, either both by overall scores (denoted as \textbf{DOVER}) or by respective aesthetic and technical opinions (denoted as \textbf{DOVER++}), discussed in Sec.~\ref{sec:objectives}. Finally, we discuss the subjectively-inspired fusion (Sec.~\ref{sec:fusion}) to predict the overall quality from DOVER.

\subsection{Methodology: Separate the Perceptual Factors}
\label{sec:viewdecomposition}

From DIVIDE-3k, we notice that aesthetic and technical perspectives in UGC-VQA are usually associated with different perceptual factors. Specifically, as illustrated in (Fig.~\ref{fig:dividebestworst}\blue{\textbf{(a)\&(b)}}), aesthetic opinions are mostly related to \textit{semantics, composition} of objects~\cite{objiaa,cadb,distilliaa}, which are typically high-level visual perceptions. In contrast, the technical quality is largely affected by low-level visual distortions, \textit{e.g.}, \textit{blurs, noises, artifacts}~\cite{dbcnn,fastvqa,discovqa,pvq,paq2piq} (Fig.~\ref{fig:dividebestworst}\green{\textbf{(e)\&(f)}}). 

The observation inspires the View Decomposition strategy that separates the video into two views: the \textbf{Aesthetic View} ($S_\mathrm{A}$) that focus on aesthetic perception, and \textbf{Technical View} ($S_\mathrm{T}$) for vice versa. With the decomposed views as inputs, two separate aesthetic ($\mathbf{M}_\mathrm{A}$) and technical branches ($\mathbf{M}_\mathrm{T}$) evaluate different perspectives separately:
\begin{equation}
   Q_\mathrm{pred, A}  = \mathbf{M}_\mathrm{A}(S_\mathrm{A});
   Q_\mathrm{pred, T} = \mathbf{M}_\mathrm{T}(S_\mathrm{T}) 
\end{equation}

Despite that most perception related to the two perspectives can be separated, a small proportion of perceptual factors are related to both perspectives, such as \textbf{brightness} related to both \green{\textit{exposure (technical)}}~\cite{qualcomm} and \rblue{\textit{lighting (aesthetic)}}~\cite{piaadataset}, or \textbf{motion blurs} (which is occasionally considered as \rblue{\textit{good aesthetics}} but typically regarded as \red{\textit{bad technical quality}}~\cite{atqa}). Thus, we don't separate these factors and keep them in both branches. Instead, we employ inductive biases (\textit{pre-training, regularization}) and specific supervision in the DIVIDE-3k to further drive the two branches' focus on corresponding perspectives, introduced as follows.


\subsection{The Aesthetic Branch}
\label{sec:aesbranch}

To help the aesthetic branch focus on the aesthetic perspective, we first pre-train the branch with Image Aethetic Assessment database AVA~\cite{avaiaa}. We then elaborate the Aesthetic View ($S_\mathrm{A}$) and additional regularization objectives.

\begin{figure}    
\centering
   \includegraphics[width=\linewidth]{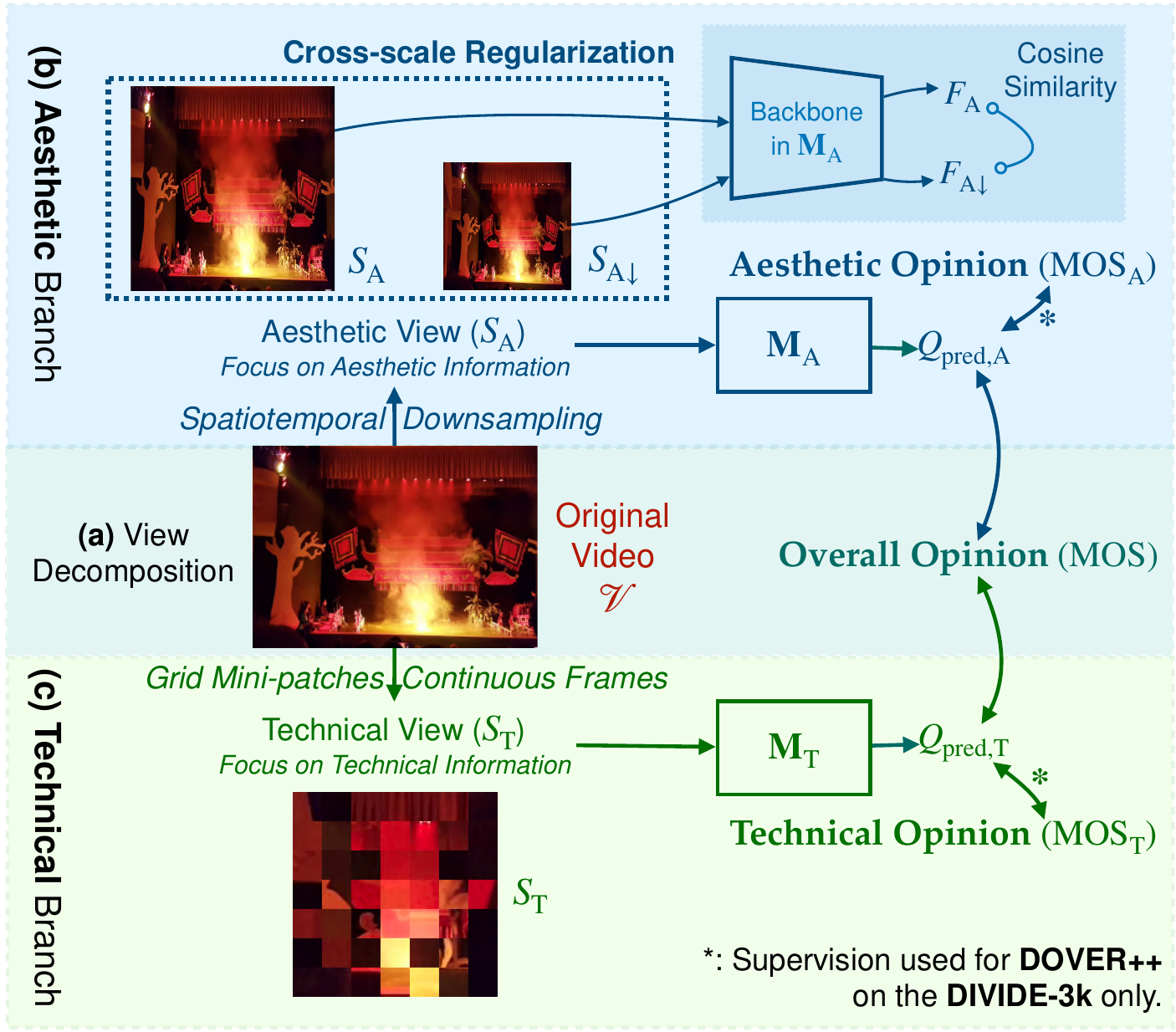}
   \vspace{-6pt}
    \caption{The proposed \underline{D}isentangled \underline{O}bjective \underline{V}ideo Quality \underline{E}valuato\underline{r} (\textbf{DOVER)} and \textbf{DOVER++} via \textbf{(a)} View Decomposition (Sec.~\ref{sec:viewdecomposition}), with the \textbf{(b)} Aesthetic Branch (Sec.~\ref{sec:aesbranch}) and the \textbf{(c)} Technical Branch (Sec.~\ref{sec:tecbranch}). The equations to obtain the two views are in Supplementary \textbf{Sec. E}.} 
    \label{fig:dover_model}
   \vspace{-16pt}
\end{figure}

\paragraph{The Aesthetic View.} \textit{Semantics} and \textit{Composition} are two key factors deciding the aesthetics of a video~\cite{aadb,cadb,distilliaa}. Thus, we obtain Aesthetic View (see Fig.~\ref{fig:dover_model}(b)) through \textit{spatial
downsampling}~\cite{bicubic} and \textit{temporal sparse frame sampling}~\cite{tsn} which preserves the semantics and composition of original videos. 
The downsampling strategies are widely applied in many existing state-of-the-art aesthetic assessment methods ~\cite{distilliaa,objiaa,racniaa,nima,gpfcnn}, further proving that they are able to preserve aesthetic information in visual contents. 
Moreover, the two strategies significantly 
{reduce the sensitivity}~\cite{videval, tlvqm, niqe, tpqi} on technical distortions such as \textit{blurs, noises, artifacts} (via spatial downsampling), \textit{shaking, flicker} (via temporal sparse sampling), so as to focus on aesthetics.

\paragraph{Cross-scale Regularization.} To better reduce technical impact in this branch, we obtain the over-downsampled view ($S_{\mathrm{A}\downarrow}$) during training by over-downsampling the videos with up to $11.3\times$ downscaling ratio. We then observe that the $S_{\mathrm{A}\downarrow}$ can barely keep any technical distortions but still remains similar aesthetics with $S_{\mathrm{A}}$ and even the original videos (see Fig.~\ref{fig:dover_model}(b) \textit{upper-right}). Furthermore, conclusions from existing study \cite{dissimilarity} suggest that feature dissimilarity among different scales (\textit{e.g.}, $S_{\mathrm{A}\downarrow}$ and $S_{\mathrm{A}}$) is related to technical distortions. Henceforth, we impose the {Cross-scale Restraint} ($\mathcal{L}_\mathrm{CR}$, Fig.~\ref{fig:dover_model}(e)) as a regularization to further reduce the technical influences in the aesthetic prediction by encouraging the feature similarity between $S_{\mathrm{A}\downarrow}$ and $S_{\mathrm{A}}$:
\begin{equation}
\mathcal{L}_\mathrm{CR} = 1 - \frac{F_{\mathrm{A}} \cdot F_{\mathrm{A}\downarrow} }{\Vert F_{\mathrm{A}} \Vert \Vert F_{\mathrm{A}\downarrow} \Vert} 
\label{eq:cr}
\end{equation}
where $F_{\mathrm{A}}$ and $F_{\mathrm{A}\downarrow}$ are output features for $S_{\mathrm{A}}$ and $S_{\mathrm{A}\downarrow}$.

\subsection{The Technical Branch}
\label{sec:tecbranch}
In the technical branch, we would like to keep the technical distortions but obfuscate the aesthetics of the videos. Thus, we design the Technical View ($S_\mathrm{T}$) as follows.

\paragraph{The Technical View.} We introduce \frag~\cite{fastvqa} (as in Fig.~\ref{fig:dover_model}(c)) as Technical View ($S_\mathrm{T}$) for the technical branch. The \frag~are composed of randomly cropped patches stitched together to retain the technical distortions. Moreover, it discarded most content and disrupted the compositional relations of the remaining, therefore severely corrupting aesthetics in videos.
Temporally, we apply \textit{continuous frame sampling} for $S_\mathrm{T}$ to retain temporal distortions.

\paragraph{Weak Global Semantics as Background.} Many studies~\cite{dbcnn,fastvqa,fastervqa} suggest that technical quality perception should consider global semantics to better assess distortion levels. Though most content is discarded in $S_\mathrm{T}$, the technical branch can still reach 68.6\% accuracy for Kinetics-400~\cite{k400data} video classification, indicating it can preserve weak global semantics as background information to distinguish textures (\textit{e.g.}, sands) from distortions (\textit{e.g.}, noises).



\subsection{Learning Objectives}
\label{sec:objectives}

\paragraph{Weak Supervision with Overall Opinions.} With the observation in Sec.~\ref{sec:3c}, the overall $\mathrm{MOS}$ can be approximated as a weighted sum of \blue{$\mathrm{MOS}_\mathrm{A}$} and \green{$\mathrm{MOS}_\mathrm{T}$}. Moreover, the subjectively-inspired inductive biases in each branch can reduce the perception of another perspective. The two observations suggest that if we use overall opinions to separately supervise the two branches, the prediction of each branch could be majorly decided by its corresponding perspective. Henceforth, we propose the Limited View Biased Supervisions ($\mathcal{L}_\mathrm{LVBS}$), which minimize the relative loss\footnote{A criterion~\cite{qaloss} based on the linear and rank correlation between predictions and labels. Details provided in supplementary \textbf{Sec.~E}.} between predictions in each branch with the overall opinion $\mathrm{MOS}$, as the objective of DOVER, applicable on all databases:
\begin{equation} 
\begin{aligned}
\mathcal{L}_\mathrm{LVBS} &= \mathcal{L}_\mathrm{Rel} (Q_\mathrm{pred, A}, \mathrm{MOS}) \\ &+ 
  \mathcal{L}_\mathrm{Rel} (Q_\mathrm{pred, T}, \mathrm{MOS}) + \lambda_\mathrm{CR}\mathcal{L}_\mathrm{CR} \\
\end{aligned}
\end{equation}

\paragraph{Supervision with Opinions from Perspectives.} With the {DIVIDE-3k} database, we further improve the accuracy for disentanglement with the Direct Supervisions ($\mathcal{L}_\mathrm{DS}$) on corresponding perspective opinions for both branches:
\begin{equation} 
\mathcal{L}_\mathrm{DS} = \mathcal{L}_\mathrm{Rel} (Q_\mathrm{pred, A}, \mathrm{MOS_A}) + 
  \mathcal{L}_\mathrm{Rel} (Q_\mathrm{pred, T}, \mathrm{MOS_T}) 
\end{equation}
and the proposed \textbf{DOVER++} is driven by a fusion of the two objectives to jointly learn more accurate overall quality as well as perspective quality predictions for each branch:
\begin{equation} 
\mathcal{L}_\mathrm{DOVER++} = \mathcal{L}_\mathrm{DS} + \lambda_\mathrm{LVBS} \mathcal{L}_\mathrm{LVBS}
\end{equation}
\subsection{Subjectively-inspired Fusion Strategy} 
\label{sec:fusion}

From the subjective studies, we observe that the $\mathrm{MOS}$ can be well-approximated as $0.428\mathrm{MOS}_\mathrm{A}+0.572\mathrm{MOS}_\mathrm{T}$. Henceforth, we propose to similarly obtain the final overall quality prediction ($Q_\mathrm{pred}$) from two perspectives: $Q_\mathrm{pred} = 0.428Q_\mathrm{pred, A} + 0.572Q_\mathrm{pred, T}$ via a simple weighted fusion. With better accuracy on all datasets (Tab.~\ref{tab:abl2branch}), the strategy by side validates the subjective observations in Sec.~\ref{sec:3c}. 



\label{sec:4d}

\section{Experimental Evaluation}

In this section, we answer two important questions:

\begin{itemize}
    \item  Can the aesthetic and technical branches better learn the effects of corresponding perspectives (Sec.~\ref{sec:evadisentangle})? 
    \item  Can the fused model more accurately predict overall quality in UGC-VQA problem (Sec.~\ref{sec:evaoverall})?
\end{itemize}   
Moreover, we include ablation studies~(Sec.~\ref{sec:abl}) and an outlook for personalized quality evaluation~(Sec.~\ref{sec:personalized}). 



\subsection{Experimental Setups}
\label{sec:impdetail}

\paragraph{Implementation Details.} 
In the aesthetic branch, we use $S_\mathrm{A}$ with size $224\times 224$ during inference and over-downsampled $S_{\mathrm{A}\downarrow}$ size $128\times128$ to better exclude technical quality issues. $N=32$ frames are sampled uniformly from each video and the backbone is inflated-ConvNext~\cite{convnext} \green{Tiny} pre-trained with AVA~\cite{avaiaa}. In the technical branch, we crop single patches at size $S_f=32$ from $7\times7$ spatial grids and sample a clip of 32 continuous frames during training, and three clips during inference. The backbone of the technical branch is Video Swin Transformer~\cite{swin3d} \green{Tiny} with GRPB~\cite{fastvqa}. $\lambda_\mathrm{CR}$ is set as $0.3$, and $\lambda_\mathrm{LVBS}$ is set as $0.5$.


\paragraph{Datasets.} Despite evaluating DOVER and DOVER++ on the proposed \textbf{DIVIDE-3k} (3,590 videos) database, we also evaluate DOVER with the large-scale UGC-VQA dataset, LSVQ~\cite{pvq} (39,072 videos), and on three smaller UGC-VQA datasets, KoNViD-1k~\cite{kv1k} (1,200 videos), LIVE-VQC~\cite{vqc} (585 videos), and YouTube-UGC~\cite{ytugccc} (1,380 videos).

\begin{table}
    \centering
    \footnotesize
    \renewcommand\arraystretch{1.2}
    \setlength\tabcolsep{8pt}
    \caption{\textbf{Quantitative Evaluation on Perspectives} of DOVER (weakly-supervised) and DOVER++ (fully-supervised) in the DIVIDE-3k, by evaluating the correlation across different predictions and subjective opinions. \textcolor{purple}{\textit{w/o} Decomposition} denotes both branches with original videos as inputs.}
    \vspace{-8pt}
    \resizebox{\linewidth}{!}{\begin{tabular}{l:c|cc}
        \hline
        Method & SROCC/PLCC & 
        \blue{$\mathrm{MOS}_\mathrm{A}$} & \green{$\mathrm{MOS}_\mathrm{T}$} \\ \hline
        \multirow{2}{79pt}{\textcolor{purple}{\textit{w/o} Decomposition} \\(\textit{w/} $\mathrm{MOS_A}$\&$\mathrm{MOS_T}$)}  &  \blue{$Q_{\mathrm{pred, A}}$}   &{0.7482/0.7576}& 0.7941/0.8039  \\
        & \green{$Q_{\mathrm{pred, T}}$}  &0.7234/0.7430& {0.8190/0.8233} \\ \hline
        \multirow{2}{75pt}{\textbf{DOVER}    \\\textcolor{purple}{\textit{w/o}} $\mathrm{MOS_A}$\&$\mathrm{MOS_T}$} &  \blue{$Q_{\mathrm{pred, A}}$}   &\underline{0.7489/0.7607}& 0.7877/0.8044  \\
        & \green{$Q_{\mathrm{pred, T}}$}  &0.7153/0.7382&\underline{0.8213/0.8295} \\ \hline
        \multirow{2}{75pt}{\textbf{DOVER++}  
  \\\textit{w/} $\mathrm{MOS_A}$\&$\mathrm{MOS_T}$} & \blue{$Q_{\mathrm{pred, A}}$} &\bred{0.7683/0.7779}&0.7584/0.7708 \\
        & \green{$Q_{\mathrm{pred, T}}$} &0.7015/0.7230&\bred{0.8376/0.8443} \\
         \hline
         
    \end{tabular}}
    \label{tab:disentanglement}
    \vspace{-9pt}
\end{table}

\begin{figure}
    \centering
    \vspace{-3pt}
    \includegraphics[width=0.98\linewidth]{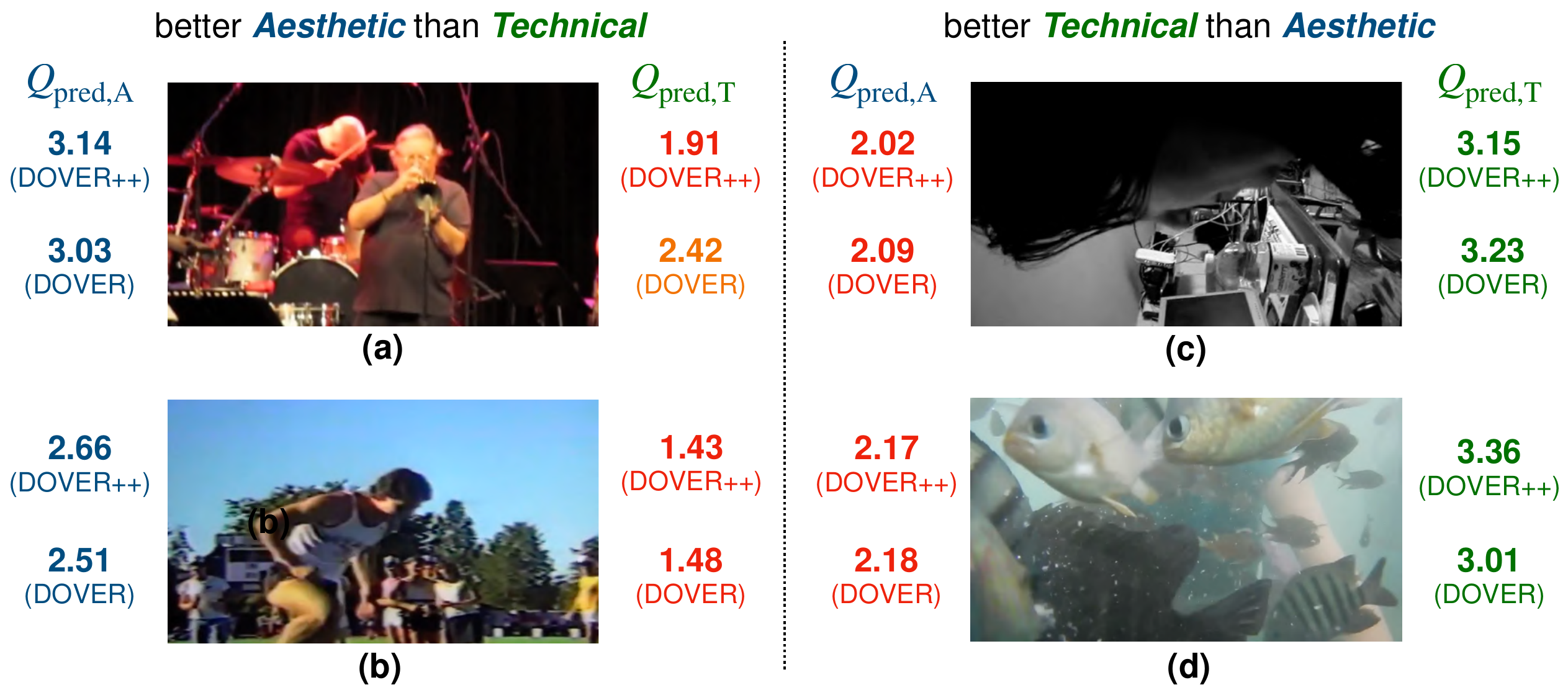}
    \vspace{-9pt}
    \caption{\textbf{Qualitative Studies on Perspectives} of DOVER/DOVER++: Visualizations of videos in the \textbf{DIVIDE-3k} where aesthetic and technical predictions are diverged. \textcolor{brown}{\textit{More visualizations in supplement. \textbf{Sec.~D}.}}}
    \vspace{-16pt}
    \label{fig:qualitativedivide}
\end{figure}

\subsection{Evaluation on Two Perspectives}
\label{sec:evadisentangle}

In this section, we quantitatively and qualitatively evaluate the perspective prediction ability of proposed methods in the \textbf{DIVIDE-3k} (Sec.~\ref{sec:60}). The divergence map and pairwise user studies further prove that the two branches in DOVER better align with human opinions on corresponding perspectives on existing UGC-VQA databases (Sec.~\ref{sec:6a}).

\subsubsection{Evaluation on the DIVIDE-3k}
\label{sec:60}

\paragraph{Quantitative Studies.} In Tab.~\ref{tab:disentanglement}, we evaluate the cross-correlation between the aesthetic and technical predictions in DOVER or DOVER++ and human opinions from the two perspectives in the DIVIDE-3k, compared with baseline (\textit{with respective labels as supervision, but without View Decomposition}). DOVER shows a stronger perspective preference than the baseline even without using the respective labels, proving the effectiveness of the decomposition strategy. DOVER++ more effectively disentangle the two perspectives with each branch around \textbf{7\%} more correlated with respective opinions than opinions from another perspective.


\paragraph{Qualitative Studies.} In Fig.~\ref{fig:qualitativedivide}, we visualize several videos with diverged predicted aesthetic and technical scores. The two videos with better aesthetic scores (Fig.~\ref{fig:qualitativedivide}\textbf{(a)\&(b)}) have clear semantics yet suffer from blurs and artifacts; on the contrary, the two with better technical scores (Fig.~\ref{fig:qualitativedivide}\textbf{(c)\&(d)}) are sharp but with chaotic composition and unclear semantics. These examples align with human perception of the two perspectives, proving that both variants can effectively provide disentangled quality predictions.


\vspace{-8pt}
\subsubsection{Evaluation on Existing UGC-VQA Datasets}
\label{sec:6a}
\begin{figure}
    \centering
    \vspace{-3pt}
    \includegraphics[width=0.93\linewidth]{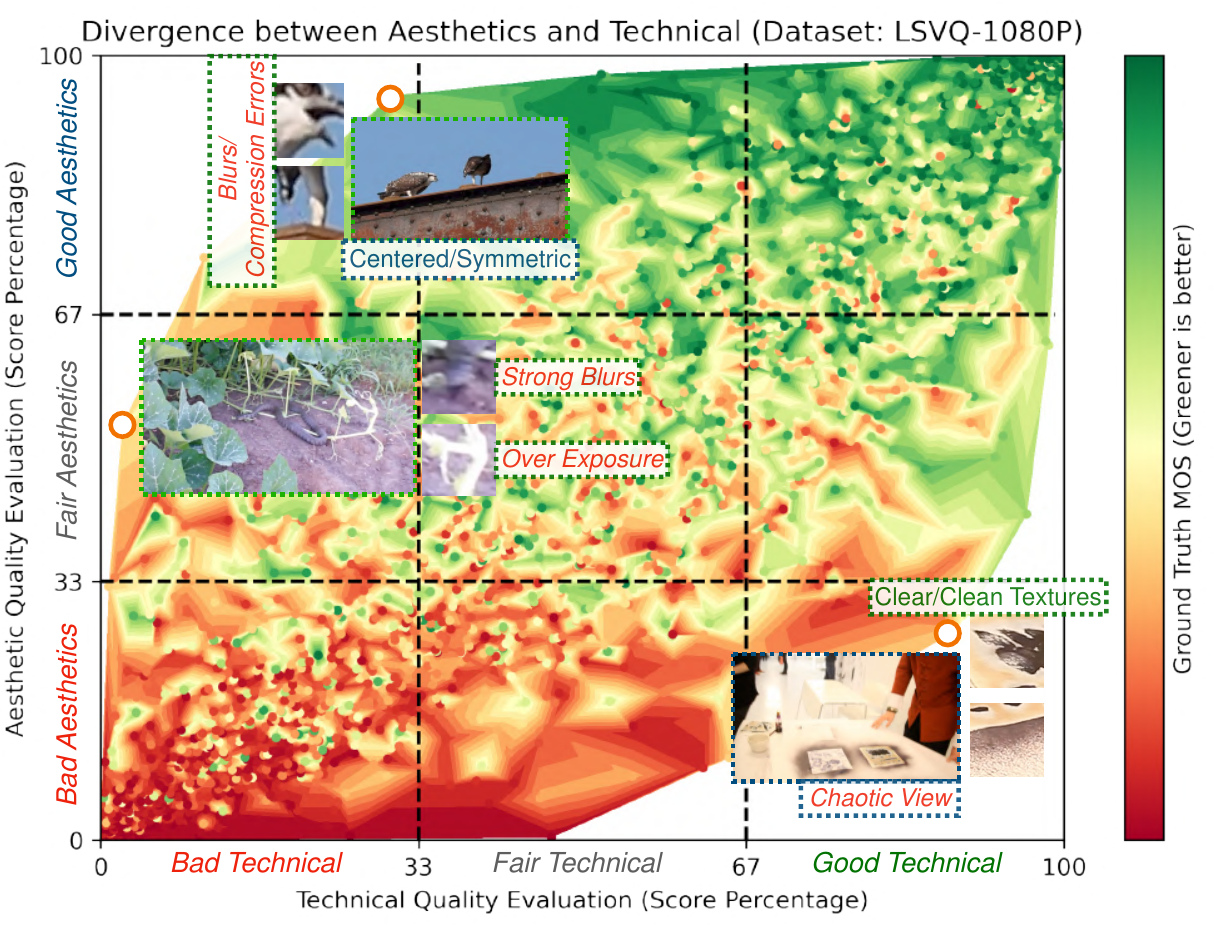}
    \vspace{-12pt}
    \caption{The divergence map of technical and aesthetic predictions of DOVER in LSVQ~\cite{pvq} dataset. Similar as Fig.~\ref{fig:qualitativedivide}, the videos with diverged scores also align with human opinions of aesthetic and technical quality.}
    \vspace{-10pt}
    \label{fig:div}
\end{figure}

\paragraph{The Divergence Map.} In Fig.~\ref{fig:div}, we visualize the divergence map between predictions in two branches (trained and tested on LSVQ~\cite{pvq}) and examine the videos where two branches score most differently, noted in \orange{\textit{orange circles}}. Among these videos, the aesthetic branch can distinguish between bad (chaotic scene, Fig.~\ref{fig:div} \textit{downright}) and {good (symmetric view, Fig.~\ref{fig:div} \textit{upleft}) aesthetics}, while the technical branch can detect technical quality issues (\textit{blurs, over-exposure, compression errors} at Fig.~\ref{fig:div} \textit{upleft}).

\begin{figure}
    \centering
    \includegraphics[width=0.91\linewidth]{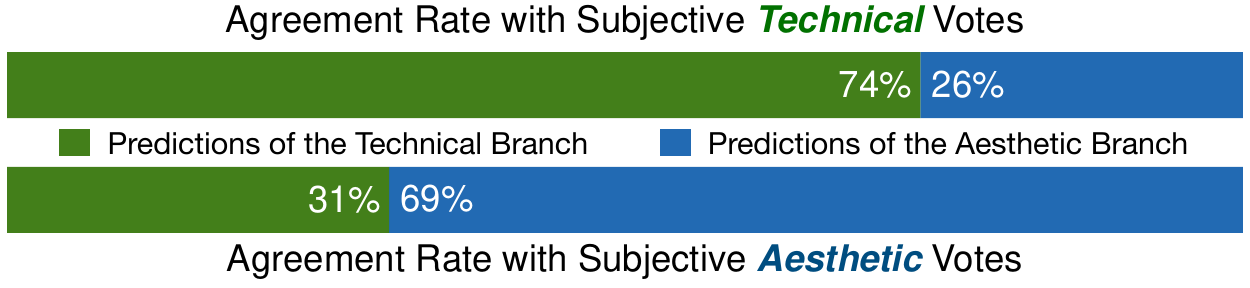}
    \vspace{-9pt}
    \caption{\textbf{User Studies on Diverged Pairs} when technical and aesthetic branches in DOVER predict differently, demonstrating that predictions of each branch are more aligned with corresponding subjective opinions.}
    \vspace{-17pt}
    \label{fig:votes}
\end{figure}

\begin{table*}[htbp]
\footnotesize
\setlength\tabcolsep{7pt}
\renewcommand\arraystretch{1.1}
\footnotesize
\center
\caption{Benchmark on official splits on the large-scale UGC-VQA dataset LSVQ~\cite{pvq}. First, second and third bests are labelled in \bred{red}, \blue{blue} and \textbf{boldface}.}\label{tab:peer}
\vspace{-8pt}
\resizebox{\linewidth}{!}{\begin{tabular}{l:ccc|cc|cc|cc|cc}
\hline
\textbf{Training Set: LSVQ$_\textbf{train}$~\cite{pvq}}& \multicolumn{3}{c|}{Inference Computational Cost} & \multicolumn{4}{c|}{Intra-dataset Evaluations}     &  \multicolumn{4}{c}{Generalization Evaluations}            \\ \hdashline
{\textbf{Testing Set}/} & \multicolumn{3}{c|}{\textit{on a 1080P, 10-second video}} & \multicolumn{2}{c|}{\textbf{LSVQ$_\text{test}$}}   & \multicolumn{2}{c|}{\textbf{LSVQ$_\text{1080p}$}}        &  \multicolumn{2}{c|}{\textbf{KoNViD-1k}}  & \multicolumn{2}{c}{\textbf{LIVE-VQC}}             \\ \hline
Methods    & GFLOPs & CPU Time & GPU Time & SROCC$\uparrow$& PLCC$\uparrow$  & SROCC$\uparrow$& PLCC$\uparrow$         & SROCC$\uparrow$& PLCC$\uparrow$                    & SROCC$\uparrow$& PLCC$\uparrow$                             \\ \hline 
\multicolumn{11}{l}{\textit{Classical Approaches (based on handcraft features):}}             \\ \hdashline

TLVQM (TIP, 2019) \cite{tlvqm} & NA & 248s & NA   & 0.772 &  0.774  & 0.589 & 0.616     & 0.732 &     0.724                   & 0.670 &  0.691 \\
VIDEVAL (TIP, 2021) \cite{videval} & NA & 895s & NA  & 0.795 &  0.783  & 0.545 & 0.554     & 0.751 &     0.741                   & 0.630 &  0.640 \\\hdashline   \multicolumn{10}{l}{\textit{Deep Approaches (based on deep neural network features):}}             \\ \hdashline 

VSFA (ACM MM, 2019) \cite{vsfa}  & 40919 & 466s & 11.1s     & 0.801 &  0.796  & 0.675 & 0.704     & 0.784 &     0.795                   & 0.734 &  0.772 \\

$\star$ Patch-VQ$_\textit{w/o patch}$ (CVPR, 2021) \cite{pvq} & 58501 & 539s & 13.8s & 0.814 &  0.816  & 0.686 &   0.708         & 0.781 &     0.781                   & 0.747 &  0.776                               \\ 
$\star$ Patch-VQ$_\textit{w/ patch}$ (CVPR, 2021) \cite{pvq}  & \multicolumn{3}{c|}{\textit{-- -- same as above -- --}}  & 0.827 &  0.828 & 0.711 &  0.739     & 0.791 &     0.795  & 0.770 &  0.807 \\ 
$\star$ {Li \textit{et al.} (TCSVT, 2022) }\cite{bvqa2021} & 112537 & 1567s & 27.6s &  {0.852} & {0.855} & {0.771} & {0.782} &{0.834} & {0.837}         &  {0.816} & {0.824} \\


{FAST-VQA (ECCV, 2022) } \cite{fastvqa} & \green{\textbf{279.1}} & \green{\textbf{8.8s}} & \green{\textbf{45ms}} & {\blue{0.876}} & {\blue{0.877}}  & {\blue{0.779}} & {\blue{0.814}} & \blue{0.859} & \blue{0.855} & {\blue{0.823}} & {\blue{0.844}}  \\ \hline


{\textbf{DOVER} (Ours)} & \green{\textbf{282.3}} & \green{\textbf{9.7s}} & \green{\textbf{47ms}}  &  {\bred{{0.888}}} & {\bred{{0.889}}}  & {\bred{{0.795}}} & {\bred{{0.830}}} & {\bred{{0.884}}} & {\bred{{0.883}}}& {\bred{{0.832}}} & {\bred{{0.855}}}  \\ \hdashline
\textit{Improvement to existing best} & -- & -- & -- & +1.3\% & +1.3\% & +2.0\% & +2.0\% & +2.9\% & +3.3\% & +1.0\% & +1.3\% \\
\hline
\end{tabular}}
\vspace{-7pt}
\end{table*}

\begin{table*}
\footnotesize
\caption{Performance benchmark on existing smaller UGC-VQA datasets. All experiments are conducted under 10 train-test splits.}\label{table:transfer}
\setlength\tabcolsep{7pt}
\renewcommand\arraystretch{1.1}
\footnotesize
\centering
\vspace{-8pt}
\resizebox{\textwidth}{!}{\begin{tabular}{l:l|cc|cc|cc|cc}
\hline
\multicolumn{2}{c|}{\textbf{Target (Fine-tuning) Quality Dataset}}       & \multicolumn{2}{c|}{\textbf{LIVE-VQC} (585)}   & \multicolumn{2}{c|}{\textbf{KoNViD-1k} (1200)}    & \multicolumn{2}{c|}{\textbf{YouTube-UGC} (1380)}   &  \multicolumn{2}{c}{\textit{Weighted Average}}        \\ \hline
\multirow{2}{120pt}{Methods} & \multirow{2}{75pt}{\textbf{Source (Pre-training) Quality Dataset}}& \multicolumn{2}{c|}{(240P - \textbf{1080P})}   & \multicolumn{2}{c|}{(540P)}      
 & \multicolumn{2}{c|}{(360P - \textbf{2160P(4K)})}         &  \\ 
     &           & SROCC$\uparrow$& PLCC$\uparrow$  & SROCC$\uparrow$& PLCC$\uparrow$         & SROCC$\uparrow$& PLCC$\uparrow$               &SROCC$\uparrow$   & PLCC$\uparrow$                             \\ \hline 

\nonecolor{lightcyan} TLVQM (TIP, 2019) \cite{tlvqm} & {NA (\textit{pure handcraft})}     & 0.799 &  0.803  & 0.773 & 0.768   & 0.669 &  0.659   & 0.732  & 0.726    \\

\nonecolor{lightcyan} VIDEVAL (TIP, 2021) \cite{videval} &   {NA (\textit{pure handcraft})}   & 0.752 &  0.751  & 0.783 & 0.780           & 0.779 &  0.773    &     0.772 & 0.772       \\ \hdashline
\nonecolor{lightcyan} RAPIQUE (OJSP, 2021) \cite{rapique} & {\textit{handcraft} + KoNiQ\cite{koniq}}       & 0.755 &  0.786  & 0.803 & 0.817                & 0.759 &  0.768    & 0.774 &     0.790 \\ 
\nonecolor{lightcyan} CNN+TLVQM (ACMMM, 2020) \cite{cnntlvqm}  &   {\textit{handcraft} + KoNiQ\cite{koniq}}   & 0.825 & 0.834 & 0.816 & 0.818  & 0.809 & 0.802 & 0.815 & 0.814  \\
\nonecolor{lightcyan} CNN+VIDEVAL (TIP, 2021) \cite{videval}   &  {\textit{handcraft} + KoNiQ\cite{koniq}}   & 0.785 & 0.810 & 0.815 & 0.817  & 0.808 & 0.803 & 0.806 & 0.810 \\
\hdashline
\nonecolor{lightcyan} VSFA (ACMMM, 2019) \cite{vsfa}  & \textit{None}        & 0.773 &  0.795  & 0.773 & 0.775   & 0.724 &  0.743   & 0.752 & 0.765 \\
\nonecolor{lightcyan} Patch-VQ (CVPR, 2021) \cite{pvq} & PaQ-2-PiQ\cite{paq2piq}  & {0.827} &  {0.837}  & 0.791 &   0.786        & NA &  NA   & NA & NA              \\
\nonecolor{lightcyan} CoINVQ (CVPR, 2021) \cite{rfugc} & \textit{self-collected} & NA &  NA & 0.767 &  0.764    & {0.816} &     {0.802} & NA & NA    \\ 
\nonecolor{lightcyan} Li \textit{et al.} (TCSVT, 2022) \cite{bvqa2021} & \textit{fused} (\cite{bid,spaq,livechallenge,koniq}) & 0.834 & 0.842 & 0.834 & 0.836 & 0.818 & 0.826 & 0.823 & 0.833  \\ 
\nonecolor{lightgray}
{{FAST-VQA} (ECCV, 2022)\cite{fastvqa}} &  LSVQ~\cite{pvq} &  {\blue{0.849}} & {\blue{0.862}} & {\blue{0.891}} & {\blue{0.892}} & {\blue{0.855}} & {\blue{0.852}}  & \blue{0.868} & \blue{0.869} \\ \hdashline
\nonecolor{lightgray} {\textbf{DOVER} (ours)} &   LSVQ~\cite{pvq} &  \textbf{\red{0.860}} & \textbf{\red{0.875}} & \textbf{\red{0.909}} & \textbf{\red{0.906}} & \textbf{\red{0.890}} & \textbf{\red{0.891}}  & \textbf{\red{0.891}} & \textbf{\red{0.891}}  \\ 
\nonecolor{lightgray} -- \textit{improvement to existing best} & & +1.6\% & +1.4\% & +2.0\% & +1.6\% & +3.9\% & +3.8\% & +2.6\% & +2.5\%  \\ 
\hline
\end{tabular}}
\vspace{-10pt}
\end{table*}

\paragraph{Pairwise User Studies.}
\label{sec:6b}
We further conduct \textbf{\textit{user studies}} to measure whether the two evaluators can distinguish the two perspectives on these diverged cases. Specifically, we evaluate on diverged pairs $\{\mathcal{V}_1, \mathcal{V}_2\}$ where aesthetic branch predicts $\mathcal{V}_1$ is obviously better (\textit{at least one score higher when scores are in the range $[1,5]$}) yet technical branch predicts $\mathcal{V}_2$ is obviously better. After random sampling 200 pairs in this way, we ask 15 subjects to choose {\textbf{which one has better {{aesthetic}} (or {{technical}}) quality in the pair}}. After post-processing the subject choices with popular votes, we calculate the agreement rates between subjective votes and predictions (in Fig.~\ref{fig:votes}). Each subjective perspective is notably more agreed with corresponding branch predictions, demonstrating that even without the respective labels, the DOVER can still learn to primarily disentangle the two perspectives. \textcolor{brown}{\textit{More details are in supplementary (\textbf{Sec.~B}).}}

\subsection{Evaluation on Overall Quality Prediction}
\label{sec:evaoverall}


\subsubsection{Results on Existing UGC-VQA Datasets}
\label{sec:benchmark}

\begin{table}
\footnotesize
\vspace{-3pt}
\caption{Performance benchmark on the \textbf{DIVIDE-3k}. All experiments are conducted under 10 train-test splits with random seed $42$. }\label{table:divide3k}
\setlength\tabcolsep{5.8pt}
\renewcommand\arraystretch{1.2}
\footnotesize
\centering
\vspace{-8pt}
\resizebox{\linewidth}{!}{\begin{tabular}{l:l|ccc}
\hline
\multicolumn{2}{c}{Training/Testing on}    & \multicolumn{3}{c}{\textbf{DIVIDE-3k} (3590)}  \\ \hline
 Methods & \textbf{Pre-training Dataset} & SROCC$\uparrow$& PLCC$\uparrow$& KROCC$\uparrow$\\ \hline
 TLVQM (2019)~\cite{tlvqm} & {NA (\textit{pure handcraft})}  & 0.6461 & 0.6807 & 0.4699 \\
 VIDEVAL (2021)~\cite{videval} & {NA (\textit{pure handcraft})} &  0.7056 & 0.7162 & 0.5233 \\ \hdashline
 RAPIQUE (2021)~\cite{rapique} & \textit{handcraft} + KoNiQ~\cite{koniq} &  0.7341 & 0.7547 & 0.5498 \\ \hdashline
 VSFA (2019)~\cite{vsfa} & NA & 0.7254 & 0.7386 & 0.5395 \\
 MDTVSFA (2021)~\cite{mdtvsfa} & NA & 0.7522 & 0.7409 & 0.5647 \\
  UNIQUE (2021)~\cite{unique} & \textit{fused} (\cite{bid,spaq,livechallenge,koniq}) & 0.7529 & 0.7637 & 0.5634 \\
 Li \textit{et al.} (2022)~\cite{bvqa2021} & \textit{fused} (\cite{bid,spaq,livechallenge,koniq}) & 0.7967 & 0.8125 & 0.6138 \\
 FAST-VQA (2022)~\cite{fastvqa} &  LSVQ~\cite{pvq} & \textbf{0.8184} & \textbf{0.8288} & \textbf{0.6285} \\ \hline
 \textbf{DOVER} (Ours) &  LSVQ~\cite{pvq} & \blue{0.8331} & \blue{0.8438} & \blue{0.6480} \\ \hdashline 
 
  \textbf{DOVER++} (Ours) &  LSVQ~\cite{pvq} & \bred{0.8442} & \bred{0.8537} & \bred{0.6603} \\ \hline 

\end{tabular}}
\vspace{-15pt}
\end{table}

\paragraph{Results on LSVQ.} In Tab.~\ref{tab:peer}, we train the DOVER on the large-scale UGC-VQA dataset, LSVQ~\cite{pvq}, and test it on five different existing UGC-VQA datasets. The proposed DOVER outperforms state-of-the-arts for intra-dataset evaluations by improving up to \bred{2.0\%} PLCC.
When testing on datasets other than LSVQ as generalization evaluation, the DOVER has shown more competitive performance. It improves PLCC on FAST-VQA by \bred{3.3\%} on KoNViD-1k, the UGC-VQA dataset with more diverse contents, further suggesting the importance of modelling from the aesthetic perspective in quality assessment on videos of diverse contents.

\begin{table}
\footnotesize
\caption{Zero-shot or cross-dataset evaluations on the \textbf{DIVIDE-3k}. None of the listed methods has been trained on the DIVIDE-3k. } 
\vspace{-8pt}
\label{table:zeroshotdivid3k}
\setlength\tabcolsep{10pt}
\renewcommand\arraystretch{1.2}
\footnotesize
\centering
\resizebox{\linewidth}{!}{\begin{tabular}{l:l|ccc}
\hline
\multicolumn{2}{c}{Evaluating on}    & \multicolumn{3}{c}{\textbf{DIVIDE-3k} (3590)}  \\ \hline
 \multicolumn{5}{l}{\textit{Zero-shot (Opinion-Unaware) VQA Approaches:}} \\ \hdashline
 Methods & \textbf{Training on} & SROCC$\uparrow$& PLCC$\uparrow$& KROCC$\uparrow$\\ \hline
 NIQE (2013)~\cite{niqe} & None & 0.3524 & 0.3839 & 0.2634 \\
TPQI (2022)~\cite{tpqi} &None & 0.4407 & 0.4432 & 0.3045 \\ 
 CLIP-IQA (2022)~\cite{clipiqa} &  CLIP~\cite{clip} & 0.5882 & 0.5910 & 0.4067 \\ 
 BVQI (2023)~\cite{buonavista} & CLIP~\cite{clip} & 0.6678 & 0.6802 & 0.4842 \\ \hdashline
  \multicolumn{5}{l}{\textit{Cross-dataset Evaluation (training on LSVQ):}} \\ \hdashline

 {Patch-VQ} (2021)~\cite{pvq} &  \multirow{3}{0pt}{LSVQ~\cite{pvq}} & 0.6454 & 0.6713 & 0.4489 \\ 
  {Li et al.} (2022)~\cite{bvqa2021} & & 0.7318 & 0.7524 & 0.5395 \\ \cdashline{1-1} \cdashline{3-5}
 \textbf{DOVER} (Ours) &   & \bred{0.7727} & \bred{0.7806} & \bred{0.5799} \\  \hline 
\end{tabular}}
\vspace{-15pt}
\end{table}

\paragraph{Results on Smaller UGC-VQA Datasets.} Following \cite{fastvqa}, we pre-train the proposed DOVER on LSVQ instead of IQA datasets~\cite{koniq,bid,paq2piq} and then fine-tune the proposed method on three smaller UGC-VQA datasets and list the results in Tab.~\ref{table:transfer}. DOVER has reached unprecedented performance on all three datasets (mean PLCC $>0.89$), and outperformed FAST-VQA with an average of \bred{2.6\%} improvement under exactly the same training process. The results further prove the effectiveness of considering aesthetic and technical perspectives separately and explicitly in UGC-VQA.

\subsubsection{Results on the DIVIDE-3k}
\label{sec:evadivide}

\paragraph{Training and Testing on DIVIDE-3k.} We first benchmark recent state-of-the-arts by conducting training and testing in the DIVIDE-3k. As shown in Tab.~\ref{table:divide3k}, the two semantic-unaware classical methods~\cite{tlvqm,videval} are performing notably worse and DOVER again achieves state-of-the-art. It is also noteworthy that with aesthetic and technical scores as auxiliary labels, DOVER++ further improves the performance for overall quality prediction. This further suggests that better modeling of the two perspectives can finally benefit overall quality assessment in the UGC-VQA problem.

\paragraph{Zero-shot and Cross-dataset Evaluations.} We also benchmark the opinion-unaware (\textit{i.e.} zero-shot) VQA approaches on the \textbf{DIVIDE-3k}. Among them, the recent BVQI~\cite{buonavista} reaches the best performance by considering both technical and semantic (aesthetic-related) criteria. Moreover, we benchmark the best approaches in Tab.~\ref{tab:peer} on the cross-dataset generalization from LSVQ to the DIVIDE-3k, where the proposed DOVER again outperforms other methods, suggesting the alignment between the proposed objective approach and subjective database.

\subsection{Ablation Studies}
\label{sec:abl}

\paragraph{Effects of View Decomposition.} In Tab.~\ref{tab:abllvbs}, we compare the proposed View Decomposition strategy with common strategies in UGC-VQA by keeping other parts the same. First of all, it is much better than the variant \textit{w/o Decomposition} that directly takes the original videos as inputs of both branches, showing the effectiveness of decomposition. Moreover, with backbone and input kept the same, DOVER with separate supervisions is also notably better than \textit{Feature Aggregation}, which first concatenates features from two branches together and then regress them to the quality scores, as applied by several existing approaches~\cite{bvqa2021,svqa,pvq}.


\begin{table}
\setlength\tabcolsep{4pt}
\renewcommand\arraystretch{1.2}
\footnotesize
\caption{\textbf{Ablation Study of DOVER (I):} the View Decomposition scheme.} 
\vspace{-9pt}
\centering
\resizebox{\linewidth}{!}{\begin{tabular}{l|c|c|c|c}
\hline
\textbf{Testing Set}/         & \multicolumn{1}{c|}{\textbf{LSVQ$_\text{test}$}}   & \multicolumn{1}{c|}{\textbf{LSVQ$_\text{1080p}$}}        &  \multicolumn{1}{c|}{\textbf{KoNViD-1k}}  & \multicolumn{1}{c}{\textbf{LIVE-VQC}}            \\ \cline{2-5}
Variants/Metric                   & SROCC/PLCC  & SROCC/PLCC         & SROCC/PLCC                    & SROCC/PLCC \\ \hline                

\textcolor{purple}{\textit{w/o} Decomposition} & 0.859/0.858 & 0.752/0.798 & 0.851/0.850 & 0.816/0.834 \\
\textit{Feature Aggregation} & 0.873/0.874 & 0.776/0.811 & 0.863/0.864 & 0.813/0.839 \\ \hline
\textbf{DOVER (Ours)} & \bred{0.888}/\bred{0.889} & \bred{0.795}/\bred{0.830} & \bred{0.884}/\bred{0.883} & \bred{0.832}/\bred{0.855}\\ 
\hline
\end{tabular}}
\label{tab:abllvbs}
\setlength\tabcolsep{4pt}
\renewcommand\arraystretch{1.17}
\footnotesize
\caption{\textbf{Ablation Study of DOVER (II):} Accuracy of single branch predictions and the effect of subjectively-inspired fusion (denoted as \textit{SIF}).} 
\vspace{-9pt}
\centering
\resizebox{\linewidth}{!}{\begin{tabular}{ccc|c|c|c|c}
\hline
\multicolumn{3}{c}{\textbf{Testing Set}/}         & \multicolumn{1}{c|}{\textbf{LSVQ$_\text{test}$}}   & \multicolumn{1}{c|}{\textbf{LSVQ$_\text{1080p}$}}        &  \multicolumn{1}{c|}{\textbf{KoNViD-1k}}  & \multicolumn{1}{c}{\textbf{LIVE-VQC}}            \\ \hline
$Q_\mathrm{pred, A}$ & $Q_\mathrm{pred, T}$ &  \textit{SIF}       & SROCC/PLCC  & SROCC/PLCC         & SROCC/PLCC                    & SROCC/PLCC\\ \hline                

\cmark & & &  0.855/0.856 & 0.738/0.782 & 0.844/0.853 & 0.792/0.826 \\
& \cmark & &  0.877/0.878 & 0.778/0.812 & 0.861/0.855 & 0.825/0.844 \\ 
\hline
\cmark & \cmark  & & 0.885/0.886 & 0.792/0.826 & 0.880/0.880 & 0.829/0.849 \\  \hdashline
\rowcolor{lightpink} \cmark & \cmark & \cmark & \bred{0.888}/\bred{0.889} & \bred{0.795}/\bred{0.830} & \bred{0.884}/\bred{0.883} & \bred{0.832}/\bred{0.855}\\ \hline
\end{tabular}}
\label{tab:abl2branch}
\vspace{-10pt}
\end{table}

\paragraph{Effects of Subjectively-Inspired Fusion.} We discuss the fusion strategy in Tab.~\ref{tab:abl2branch}. As shown in the table, only considering one branch will bring a notable performance decrease, and directly obtaining the fused quality as $Q_\mathrm{pred, A} + Q_\mathrm{pred, T}$ without weights is also less accurate than subjectively-inspired fusion. These results further validate the subjective observations found in the DIVIDE-3k.

\paragraph{Ablation Studies of DOVER++.} In Tab.~\ref{tab:abldoverpp}, we further discuss whether the extra objective ($\mathcal{L}_\mathrm{DS}$) can improve accuracy of overall quality prediction. By combining $\mathcal{L}_\mathrm{DS}$ with $\mathcal{L}_\mathrm{LVBS}$, it contributes to around 1\% performance gain. It is also noteworthy that even without direct $\mathrm{MOS}$ labels for supervision, the $\mathcal{L}_\mathrm{DS}$ only can still outperform $\mathcal{L}_\mathrm{LVBS}$. All these results suggest that explicitly considering ``quality" in UGC-VQA into a sum of two perspectives is a good approximation to the human perceptual mechanism.

\subsection{Outlook: Personalized Quality Evaluation}
\label{sec:personalized}

\begin{figure}
    \centering
    \includegraphics[width=0.99\linewidth]{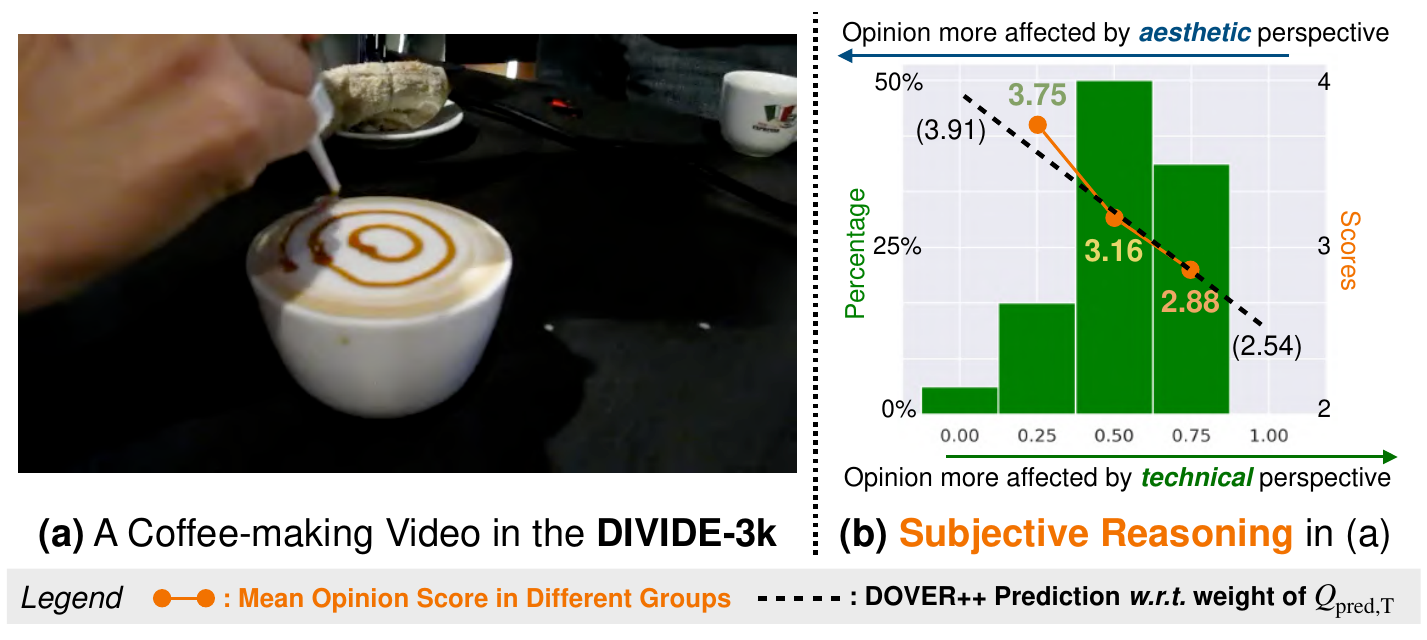}
    \vspace{-8pt}
    \caption{For the video \textbf{(a)}, the impact of aesthetic and technical perspectives on the final quality rating \textbf{(b) varies among individuals}. By adjusting fusion weights, DOVER++ can align with opinions from different groups.}\label{fig:variantreasoning}
    \vspace{-15pt}
\end{figure}

 During the subjective reasoning study, we further find out that the effect of each perspective varies among different individuals. For instance, the video in Fig.~\ref{fig:variantreasoning}{(a)} has {\rblue{better aesthetics}} and \red{worse technical quality} (\textit{blurry, under-exposed}), and different individuals consider the technical impact differently while rating the overall opinion (Fig.~\ref{fig:variantreasoning}{(b)}).
 Moreover, with more consideration of the technical perspective, subjects tend to rate lower scores on the video. With DOVER++, if we adaptively fuse between $Q_\mathrm{pred,A}$ and $Q_\mathrm{pred,T}$, we find that the differently-fused results can better predict the quality perception of individual subject groups, suggesting its primary capability to provide quality evaluation catering for personalized requirements.

\begin{table}
\setlength\tabcolsep{7pt}
\renewcommand\arraystretch{1.1}
\footnotesize
\caption{\textbf{Ablation Study of DOVER++}: Effects of different objectives.} 
\vspace{-9pt}
\centering
\resizebox{\linewidth}{!}{
\begin{tabular}{l:cc|ccc}
\hline
 \multicolumn{3}{c|}{Loss Objectives}                 & \multicolumn{3}{c}{\textbf{DIVIDE-3k} (3590)} \\ \hline
  & $\mathcal{L}_\mathrm{LVBS}$ & $\mathcal{L}_\mathrm{DS}$ & SROCC$\uparrow$       & PLCC$\uparrow$ & KROCC$\uparrow$ \\
\hline
\textit{w/o} $\mathrm{MOS_A}$\&$\mathrm{MOS_T}$   & \cmark  &   & 0.8331 & 0.8438 & 0.6480 \\ \hdashline
\multirow{2}{75pt}{\textit{w/} $\mathrm{MOS_A}$\&$\mathrm{MOS_T}$}   &   & \cmark  & \blue{0.8357} & \blue{0.8455} & \blue{0.6521} \\ 
 & \cellcolor{lightpink} \cmark & \cellcolor{lightpink} \cmark & \cellcolor{lightpink}\bred{0.8442} & \cellcolor{lightpink}\bred{0.8537} & \cellcolor{lightpink}\bred{0.6603} \\
\hline
\end{tabular}}
\label{tab:abldoverpp}
\vspace{-10pt}
\end{table}

\section{Conclusion}

In this paper, we present the DIVIDE-3k database and the first subjective study aimed at exploring the impact of aesthetic and technical perspectives on UGC-VQA, which reveals that both perspectives impact human quality opinions. In light of this observation, we propose the objective quality evaluators, DOVER and DOVER++, that achieve two objectives: \textbf{1)} significantly improving overall UGC-VQA performance; \textbf{2)} decoupling effects of two perspectives, so as to be applicable to specific real-world scenarios where pure technical or aesthetic quality metrics are needed.

\clearpage
{\small
\bibliographystyle{IEEEtran}
\bibliography{egbib}

\begin{thebibliography}{10}
\providecommand{\url}[1]{#1}
\csname url@samestyle\endcsname
\providecommand{\newblock}{\relax}
\providecommand{\bibinfo}[2]{#2}
\providecommand{\BIBentrySTDinterwordspacing}{\spaceskip=0pt\relax}
\providecommand{\BIBentryALTinterwordstretchfactor}{4}
\providecommand{\BIBentryALTinterwordspacing}{\spaceskip=\fontdimen2\font plus
\BIBentryALTinterwordstretchfactor\fontdimen3\font minus
  \fontdimen4\font\relax}
\providecommand{\BIBforeignlanguage}[2]{{%
\expandafter\ifx\csname l@#1\endcsname\relax
\typeout{** WARNING: IEEEtran.bst: No hyphenation pattern has been}%
\typeout{** loaded for the language `#1'. Using the pattern for}%
\typeout{** the default language instead.}%
\else
\language=\csname l@#1\endcsname
\fi
#2}}
\providecommand{\BIBdecl}{\relax}
\BIBdecl

\bibitem{pvq}
Z.~Ying, M.~Mandal, D.~Ghadiyaram, and A.~Bovik, ``Patch-vq: 'patching up' the
  video quality problem,'' in \emph{CVPR}, June 2021, pp. 14\,019--14\,029.

\bibitem{mlsp}
F.~G\"otz-Hahn, V.~Hosu, H.~Lin, and D.~Saupe, ``Konvid-150k: A dataset for
  no-reference video quality assessment of videos in-the-wild,'' in \emph{IEEE
  Access 9}.\hskip 1em plus 0.5em minus 0.4em\relax IEEE, 2021, pp.
  72\,139--72\,160.

\bibitem{kv1k}
V.~Hosu, F.~Hahn, M.~Jenadeleh, H.~Lin, H.~Men, T.~Szirányi, S.~Li, and
  D.~Saupe, ``The konstanz natural video database (konvid-1k),'' in
  \emph{QoMEX}, 2017, pp. 1--6.

\bibitem{vqc}
Z.~Sinno and A.~C. Bovik, ``Large-scale study of perceptual video quality,''
  \emph{IEEE Transactions on Image Processing}, vol.~28, no.~2, pp. 612--627,
  2019.

\bibitem{ytugccc}
Y.~Wang, S.~Inguva, and B.~Adsumilli, ``Youtube ugc dataset for video
  compression research,'' in \emph{2019 IEEE 21st International Workshop on
  Multimedia Signal Processing (MMSP)}, 2019, pp. 1--5.

\bibitem{vsfa}
D.~Li, T.~Jiang, and M.~Jiang, ``Quality assessment of in-the-wild videos,'' in
  \emph{ACM MM}, 2019, p. 2351–2359.

\bibitem{rfugc}
Y.~Wang, J.~Ke, H.~Talebi, J.~G. Yim, N.~Birkbeck, B.~Adsumilli, P.~Milanfar,
  and F.~Yang, ``Rich features for perceptual quality assessment of ugc
  videos,'' in \emph{CVPR}, June 2021, pp. 13\,435--13\,444.

\bibitem{internetvqa}
J.~Xu, J.~Li, X.~Zhou, W.~Zhou, B.~Wang, and Z.~Chen, ``Perceptual quality
  assessment of internet videos,'' in \emph{ACM MM}, 2021.

\bibitem{videval}
Z.~Tu, Y.~Wang, N.~Birkbeck, B.~Adsumilli, and A.~C. Bovik, ``Ugc-vqa:
  Benchmarking blind video quality assessment for user generated content,''
  \emph{IEEE Transactions on Image Processing}, vol.~30, pp. 4449--4464, 2021.

\bibitem{tlvqm}
J.~Korhonen, ``Two-level approach for no-reference consumer video quality
  assessment,'' \emph{IEEE Transactions on Image Processing}, vol.~28, no.~12,
  pp. 5923--5938, 2019.

\bibitem{niqe}
A.~Mittal, R.~Soundararajan, and A.~C. Bovik, ``Making a “completely blind”
  image quality analyzer,'' \emph{IEEE Signal Processing Letters}, vol.~20,
  no.~3, pp. 209--212, 2013.

\bibitem{tpqi}
L.~Liao, K.~Xu, H.~Wu, C.~Chen, W.~Sun, Q.~Yan, and W.~Lin, ``Exploring the
  effectiveness of video perceptual representation in blind video quality
  assessment,'' in \emph{ACM MM}, 2022.

\bibitem{fastvqa}
H.~Wu, C.~Chen, J.~Hou, L.~Liao, A.~Wang, W.~Sun, Q.~Yan, and W.~Lin,
  ``Fast-vqa: Efficient end-to-end video quality assessment with fragment
  sampling,'' in \emph{ECCV}, 2022.

\bibitem{dxomark}
\BIBentryALTinterwordspacing
DxOMark, ``Dxomark photography benchmark.'' [Online]. Available:
  \url{http://dxomark.com/}
\BIBentrySTDinterwordspacing

\bibitem{spaq}
Y.~Fang, H.~Zhu, Y.~Zeng, K.~Ma, and Z.~Wang, ``Perceptual quality assessment
  of smartphone photography,'' in \emph{CVPR}, 2020, pp. 3677--3686.

\bibitem{basicvsr}
K.~C. Chan, X.~Wang, K.~Yu, C.~Dong, and C.~C. Loy, ``Basicvsr: The search for
  essential components in video super-resolution and beyond,'' in \emph{CVPR},
  2021.

\bibitem{swinir}
J.~Liang, J.~Cao, G.~Sun, K.~Zhang, L.~Van~Gool, and R.~Timofte, ``Swinir:
  Image restoration using swin transformer,'' in \emph{ICCV Workshops}, 2021.

\bibitem{mmp}
Y.~Wang, Y.~Lu, Y.~Gao, L.~Wang, Z.~Zhong, Y.~Zheng, and A.~Yamashita,
  ``Efficient video deblurring guided by motion magnitude,'' in
  \emph{Proceedings of the European Conference on Computer Vision (ECCV)},
  2022.

\bibitem{h264}
T.~Wiegand, ``Draft itu-t recommendation and final draft international standard
  of joint video specification,'' 2003.

\bibitem{sfa}
D.~Li, T.~Jiang, W.~Lin, and M.~Jiang, ``Which has better visual quality: The
  clear blue sky or a blurry animal?'' \emph{IEEE Transactions on Multimedia},
  vol.~21, no.~5, pp. 1221--1234, 2019.

\bibitem{discovqa}
H.~Wu, C.~Chen, L.~Liao, J.~Hou, W.~Sun, Q.~Yan, and W.~Lin, ``Discovqa:
  Temporal distortion-content transformers for video quality assessment,''
  \emph{IEEE Transactions on Circuits and Systems for Video Technology}, 2023.

\bibitem{avaiaa}
N.~Murray, L.~Marchesotti, and F.~Perronnin, ``Ava: A large-scale database for
  aesthetic visual analysis,'' in \emph{CVPR}, 2012, pp. 2408--2415.

\bibitem{mlspiaa}
V.~Hosu, B.~Goldlücke, and D.~Saupe, ``Effective aesthetics prediction with
  multi-level spatially pooled features,'' in \emph{CVPR}, 2019, pp.
  9367--9375.

\bibitem{distilliaa}
J.~Hou, H.~Ding, W.~Lin, W.~Liu, and Y.~Fang, ``Distilling knowledge from
  object classification to aesthetics assessment,'' \emph{IEEE Transactions on
  Circuits and Systems for Video Technology}, 2022.

\bibitem{racniaa}
X.~Zhang, X.~Gao, W.~Lu, L.~He, and J.~Li, ``Beyond vision: A multimodal
  recurrent attention convolutional neural network for unified image aesthetic
  prediction tasks,'' \emph{IEEE Transactions on Multimedia}, vol.~23, pp.
  611--623, 2021.

\bibitem{piaadataset}
Y.~Yang, L.~Xu, L.~Li, N.~Qie, Y.~Li, P.~Zhang, and Y.~Guo, ``Personalized
  image aesthetics assessment with rich attributes,'' in \emph{CVPR}, 2022, pp.
  19\,861--19\,869.

\bibitem{objiaa}
J.~Hou, S.~Yang, and W.~Lin, ``Object-level attention for aesthetic rating
  distribution prediction,'' in \emph{ACM MM}, 2020, p. 816–824.

\bibitem{cvd}
M.~Nuutinen, T.~Virtanen, M.~Vaahteranoksa, T.~Vuori, P.~Oittinen, and
  J.~Häkkinen, ``Cvd2014—a database for evaluating no-reference video
  quality assessment algorithms,'' \emph{IEEE Transactions on Image
  Processing}, vol.~25, no.~7, pp. 3073--3086, 2016.

\bibitem{qualcomm}
D.~Ghadiyaram, J.~Pan, A.~C. Bovik, A.~K. Moorthy, P.~Panda, and K.-C. Yang,
  ``In-capture mobile video distortions: A study of subjective behavior and
  objective algorithms,'' \emph{IEEE Transactions on Circuits and Systems for
  Video Technology}, vol.~28, no.~9, pp. 2061--2077, 2018.

\bibitem{ytugc}
J.~G. Yim, Y.~Wang, N.~Birkbeck, and B.~Adsumilli, ``Subjective quality
  assessment for youtube ugc dataset,'' in \emph{ICIP}, 2020, pp. 131--135.

\bibitem{livevqa}
K.~Seshadrinathan, R.~Soundararajan, A.~C. Bovik, and L.~K. Cormack, ``Study of
  subjective and objective quality assessment of video,'' \emph{IEEE
  Transactions on Image Processing}, vol.~19, no.~6, pp. 1427--1441, 2010.

\bibitem{csiqvqa}
P.~V. Vu and D.~M. Chandler, ``Vis3: an algorithm for video quality assessment
  via analysis of spatial and spatiotemporal slices,'' \emph{Journal of
  Electronic Imaging}, vol.~23, 2014.

\bibitem{yfcc}
B.~Thomee, D.~A. Shamma, G.~Friedland, B.~Elizalde, K.~Ni, D.~Poland, D.~Borth,
  and L.-J. Li, ``Yfcc100m: The new data in multimedia research,''
  \emph{Commun. ACM}, vol.~59, no.~2, p. 64–73, 2016.

\bibitem{crowdsource}
D.~Ghadiyaram and A.~C. Bovik, ``Massive online crowdsourced study of
  subjective and objective picture quality,'' \emph{IEEE Transactions on Image
  Processing}, vol.~25, no.~1, pp. 372--387, 2016.

\bibitem{bofqa}
------, ``Perceptual quality prediction on authentically distorted images using
  a bag of features approach,'' \emph{Journal of Vision}, vol.~17, 2017.

\bibitem{rrstedqa}
R.~Soundararajan and A.~C. Bovik, ``Video quality assessment by reduced
  reference spatio-temporal entropic differencing,'' \emph{IEEE Transactions on
  Circuits and Systems for Video Technology}, vol.~23, pp. 684--694, 2013.

\bibitem{diivine}
A.~K. Moorthy and A.~C. Bovik, ``Blind image quality assessment: From natural
  scene statistics to perceptual quality,'' \emph{IEEE Transactions on Image
  Processing}, vol.~20, pp. 3350--3364, 2011.

\bibitem{stgreed}
P.~C. Madhusudana, N.~Birkbeck, Y.~Wang, B.~Adsumilli, and A.~C. Bovik,
  ``{ST-GREED}: Space-time generalized entropic differences for frame rate
  dependent video quality prediction,'' \emph{IEEE Trans. Image Process.},
  2021.

\bibitem{vmaf}
Z.~Li, A.~Aaron, I.~Katsavounidis, A.~Moorthy, and M.~Manohara, ``Toward a
  practical perceptual video quality metric,'' \emph{The Netflix Tech Blog},
  vol.~6, no.~2, 2016.

\bibitem{brisque}
A.~Mittal, A.~K. Moorthy, and A.~C. Bovik, ``No-reference image quality
  assessment in the spatial domain,'' \emph{IEEE Transactions on Image
  Processing}, vol.~21, no.~12, pp. 4695--4708, 2012.

\bibitem{viideo}
A.~Mittal, M.~A. Saad, and A.~C. Bovik, ``A completely blind video integrity
  oracle,'' \emph{IEEE Transactions on Image Processing}, vol.~25, no.~1, pp.
  289--300, 2016.

\bibitem{vbliinds}
M.~A. Saad, A.~C. Bovik, and C.~Charrier, ``Blind image quality assessment: A
  natural scene statistics approach in the dct domain,'' \emph{IEEE
  Transactions on Image Processing}, vol.~21, no.~8, pp. 3339--3352, 2012.

\bibitem{dctqa}
Y.~Zhang, X.~Gao, L.~He, W.~Lu, and R.~He, ``Blind video quality assessment
  with weakly supervised learning and resampling strategy,'' \emph{IEEE
  Transactions on Circuits and Systems for Video Technology}, vol.~29, pp.
  2244--2255, 2019.

\bibitem{cnn+lstm}
J.~You and J.~Korhonen, ``Deep neural networks for no-reference video quality
  assessment,'' in \emph{ICIP}, 2019, pp. 2349--2353.

\bibitem{deepvqa}
W.~Kim, J.~Kim, S.~Ahn, J.~Kim, and S.~Lee, ``Deep video quality assessor: From
  spatio-temporal visual sensitivity to a convolutional neural aggregation
  network,'' in \emph{ECCV}, 2018.

\bibitem{gstvqa}
B.~Chen, L.~Zhu, G.~Li, F.~Lu, H.~Fan, and S.~Wang, ``Learning generalized
  spatial-temporal deep feature representation for no-reference video quality
  assessment,'' \emph{IEEE Transactions on Circuits and Systems for Video
  Technology}, 2021.

\bibitem{rirnet}
P.~Chen, L.~Li, L.~Ma, J.~Wu, and G.~Shi, ``Rirnet: Recurrent-in-recurrent
  network for video quality assessment,'' \emph{ACM MM}, 2020.

\bibitem{dstsvqa}
Y.~Liu, X.~Zhou, H.~Yin, H.~Wang, and C.~C. Yan, ``Efficient video quality
  assessment with deeper spatiotemporal feature extraction and integration,''
  \emph{Journal of Electronic Imaging}, vol.~30, pp. 063\,034 -- 063\,034,
  2021.

\bibitem{svqa}
W.~Sun, X.~Min, W.~Lu, and G.~Zhai, ``A deep learning based no-reference
  quality assessment model for ugc videos,'' \emph{arXiv preprint
  arXiv:2204.14047}, 2022.

\bibitem{mdtvsfa}
D.~Li, T.~Jiang, and M.~Jiang, ``Unified quality assessment of in-the-wild
  videos with mixed datasets training,'' \emph{International Journal of
  Computer Vision}, vol. 129, no.~4, pp. 1238--1257, 2021.

\bibitem{he2016residual}
K.~He, X.~Zhang, S.~Ren, and J.~Sun, ``Deep residual learning for image
  recognition,'' in \emph{CVPR}, 2016, pp. 770--778.

\bibitem{lsctphiq}
J.~You, ``Long short-term convolutional transformer for no-reference video
  quality assessment,'' in \emph{ACM MM}, 2021, p. 2112–2120.

\bibitem{cnntlvqm}
J.~Korhonen, Y.~Su, and J.~You, ``Blind natural video quality prediction via
  statistical temporal features and deep spatial features,'' in \emph{ACM MM},
  2020, p. 3311–3319.

\bibitem{k400data}
W.~Kay, J.~Carreira, K.~Simonyan, B.~Zhang, C.~Hillier, S.~Vijayanarasimhan,
  F.~Viola, T.~Green, T.~Back, A.~Natsev, M.~Suleyman, and A.~Zisserman, ``The
  kinetics human action video dataset,'' \emph{ArXiv}, vol. abs/1705.06950,
  2017.

\bibitem{clipiqa}
J.~Wang, K.~C.~K. Chan, and C.~C. Loy, ``Exploring clip for assessing the look
  and feel of images,'' 2022.

\bibitem{matchhistogram}
V.~Vonikakis, R.~Subramanian, J.~Arnfred, and S.~Winkler, ``A probabilistic
  approach to people-centric photo selection and sequencing,'' \emph{IEEE
  Transactions on Multimedia}, vol.~19, no.~11, pp. 2609--2624, 2017.

\bibitem{bvqa2021}
B.~Li, W.~Zhang, M.~Tian, G.~Zhai, and X.~Wang, ``Blindly assess quality of
  in-the-wild videos via quality-aware pre-training and motion perception,''
  \emph{IEEE Transactions on Circuits and Systems for Video Technology}, 2022.

\bibitem{cadb}
B.~Zhang, L.~Niu, and L.~Zhang, ``Image composition assessment with
  saliency-augmented multi-pattern pooling,'' \emph{arXiv preprint
  arXiv:2104.03133}, 2021.

\bibitem{dbcnn}
W.~Zhang, K.~Ma, J.~Yan, D.~Deng, and Z.~Wang, ``Blind image quality assessment
  using a deep bilinear convolutional neural network,'' \emph{IEEE Transactions
  on Circuits and Systems for Video Technology}, vol.~30, no.~1, pp. 36--47,
  2020.

\bibitem{paq2piq}
Z.~Ying, H.~Niu, P.~Gupta, D.~Mahajan, D.~Ghadiyaram, and A.~Bovik, ``From
  patches to pictures (paq-2-piq): Mapping the perceptual space of picture
  quality,'' in \emph{CVPR}, 2020.

\bibitem{atqa}
\emph{ATQAM/MAST'20: Joint Workshop on Aesthetic and Technical Quality
  Assessment of Multimedia and Media Analytics for Societal Trends}.\hskip 1em
  plus 0.5em minus 0.4em\relax New York, NY, USA: Association for Computing
  Machinery, 2020.

\bibitem{aadb}
S.~Kong, X.~Shen, Z.~Lin, R.~Mech, and C.~Fowlkes, ``Photo aesthetics ranking
  network with attributes and content adaptation,'' in \emph{ECCV}, 2016.

\bibitem{bicubic}
R.~Keys, ``Cubic convolution interpolation for digital image processing,''
  \emph{IEEE Transactions on Acoustics, Speech, and Signal Processing},
  vol.~29, no.~6, pp. 1153--1160, 1981.

\bibitem{tsn}
L.~Wang, Y.~Xiong, Z.~Wang, Y.~Qiao, D.~Lin, X.~Tang, and L.~Van~Gool,
  ``Temporal segment networks for action recognition in videos,'' \emph{IEEE
  Transactions on Pattern Analysis and Machine Intelligence}, vol.~41, no.~11,
  pp. 2740--2755, 2019.

\bibitem{nima}
H.~Talebi and P.~Milanfar, ``Nima: Neural image assessment,'' \emph{IEEE
  Transactions on Image Processing}, vol.~27, no.~8, pp. 3998--4011, 2018.

\bibitem{gpfcnn}
X.~Zhang, X.~Gao, W.~Lu, and L.~He, ``A gated peripheral-foveal convolutional
  neural network for unified image aesthetic prediction,'' \emph{IEEE
  Transactions on Multimedia}, vol.~PP, pp. 1--1, 04 2019.

\bibitem{dissimilarity}
V.~Bhateja, A.~Kalsi, and A.~Srivastava, ``Reduced reference iqa based on
  structural dissimilarity,'' in \emph{2014 International Conference on Signal
  Processing and Integrated Networks (SPIN)}, 2014, pp. 63--68.

\bibitem{fastervqa}
H.~Wu, C.~Chen, L.~Liao, J.~Hou, W.~Sun, Q.~Yan, J.~Gu, and W.~Lin,
  ``Neighbourhood representative sampling for efficient end-to-end video
  quality assessment,'' \emph{arXiv preprint arXiv:2210.05357}, 2022.

\bibitem{qaloss}
D.~Li, T.~Jiang, and M.~Jiang, ``Norm-in-norm loss with faster convergence and
  better performance for image quality assessment,'' in \emph{ACM MM}, 2020, p.
  789–797.

\bibitem{convnext}
Z.~Liu, H.~Mao, C.-Y. Wu, C.~Feichtenhofer, T.~Darrell, and S.~Xie, ``A convnet
  for the 2020s,'' in \emph{CVPR}, 2022, pp. 11\,976--11\,986.

\bibitem{swin3d}
Z.~Liu, J.~Ning, Y.~Cao, Y.~Wei, Z.~Zhang, S.~Lin, and H.~Hu, ``Video swin
  transformer,'' in \emph{CVPR}, 2022.

\bibitem{rapique}
Z.~Tu, X.~Yu, Y.~Wang, N.~Birkbeck, B.~Adsumilli, and A.~C. Bovik, ``Rapique:
  Rapid and accurate video quality prediction of user generated content,''
  \emph{IEEE Open Journal of Signal Processing}, vol.~2, pp. 425--440, 2021.

\bibitem{koniq}
V.~Hosu, H.~Lin, T.~Sziranyi, and D.~Saupe, ``Koniq-10k: An ecologically valid
  database for deep learning of blind image quality assessment,'' \emph{IEEE
  Transactions on Image Processing}, vol.~29, pp. 4041--4056, 2020.

\bibitem{bid}
A.~Ciancio, A.~L. N.~T. Targino~da Costa, E.~A.~B. da~Silva, A.~Said,
  R.~Samadani, and P.~Obrador, ``No-reference blur assessment of digital
  pictures based on multifeature classifiers,'' \emph{IEEE Transactions on
  Image Processing}, vol.~20, no.~1, pp. 64--75, 2011.

\bibitem{livechallenge}
D.~Ghadiyaram and A.~C. Bovik, ``Massive online crowdsourced study of
  subjective and objective picture quality,'' \emph{IEEE Transactions on Image
  Processing}, vol.~25, no.~1, pp. 372--387, 2015.

\bibitem{unique}
W.~Zhang, K.~Ma, G.~Zhai, and X.~Yang, ``Uncertainty-aware blind image quality
  assessment in the laboratory and wild,'' \emph{IEEE Transactions on Image
  Processing}, vol.~30, pp. 3474--3486, Mar. 2021.

\bibitem{clip}
A.~Radford, J.~W. Kim, C.~Hallacy, A.~Ramesh, G.~Goh, S.~Agarwal, G.~Sastry,
  A.~Askell, P.~Mishkin, J.~Clark, G.~Krueger, and I.~Sutskever, ``Learning
  transferable visual models from natural language supervision,'' 2021.

\bibitem{buonavista}
H.~Wu, L.~Liao, J.~Hou, C.~Chen, E.~Zhang, A.~Wang, W.~Sun, Q.~Yan, and W.~Lin,
  ``Exploring opinion-unaware video quality assessment with semantic affinity
  criterion,'' \emph{arXiv preprint arXiv:2302.13269}, 2023.

\end{thebibliography}
}

\end{document}